\documentclass{article}

\usepackage[final, nonatbib]{bdl_2021_camera_ready}

\usepackage[utf8]{inputenc} 
\usepackage[T1]{fontenc}    
\usepackage{hyperref}       
\hypersetup{colorlinks}

\usepackage{url}            
\usepackage{booktabs}       
\usepackage{amsfonts}       
\usepackage{nicefrac}       
\usepackage{microtype}      
\usepackage{xcolor}      
\usepackage{graphicx}
\usepackage{float}
\usepackage{caption}
\usepackage{subcaption}
\usepackage{multirow}
\usepackage{placeins}

\title{Benchmark for Out-of-Distribution Detection in Deep Reinforcement Learning}

\author{%
    \parbox{\linewidth}{
        Aaqib Parvez Mohammed$^1$, Matias Valdenegro-Toro$^2$\\
    }\\
    ~\\
    \parbox{\linewidth}{
        $^1$ Bonn-Rhein-Sieg University of Applied Sciences, 53757 Sankt Augustin, Germany.\\
        $^2$ German Research Center for Artificial Intelligence, 28359 Bremen, Germany.\\
        ~\\
        \texttt{aaqib.mohammed@smail.inf.h-brs.de}, \, \texttt{matias.valdenegro@dfki.de}
    }
}

\begin{document}

\maketitle

\begin{abstract}
Reinforcement Learning (RL) based solutions are being adopted in a variety of domains including robotics, health care and industrial automation. Most focus is given to when these solutions work well, but they fail when presented with out of distribution inputs.
RL policies share the same faults as most machine learning models. Out of distribution detection for RL is generally not well covered in the literature, and there is a lack of benchmarks for this task.
In this work we propose a benchmark to evaluate OOD detection methods in a Reinforcement Learning setting, by modifying the physical parameters of non-visual standard environments or corrupting the state observation for visual environments. We discuss ways to generate custom RL environments that can produce OOD data, and evaluate three uncertainty methods for the OOD detection task. Our results show that ensemble methods have the best OOD detection performance with a lower standard deviation across multiple environments.
\end{abstract}

\section{Introduction}
Reinforcement learning (RL) is one of the paradigms of machine learning. It involves training an agent to solve tasks by interacting with the environment and learning from its experience. In the recent years, RL has been successful in a variety of domains including robotics \cite{hwangbo2019learning}, game playing \cite{silver2018general} and even agricultural applications \cite{chen2021reinforcement}.  Researchers are using it on a daily basis to make important decisions. The performance of trained agents is highly dependent on the experience or data seen during training. It is usually assumed that the test data follows the same distribution as the training data. However, this assumption does not hold true in many real world applications. The samples or observations that do not conform to the underlying distribution of the training data are referred to as out-of-distribution (OOD) samples.

Recent RL algorithms are making use of deep neural networks which are known to be sensitive to OOD data \cite{hendrycks2020many}. This can result in incorrect decisions which in turn can have significant costs. When developing new algorithms, researchers usually focus on the performance of the models calculated by metrics like accuracy and mean squared error and often ignore to report the models' uncertainty in its predictions. The uncertainty of the model can be directly associated with the trust in its predictions. Predictions with higher uncertainty can be rejected or can be an indication for the need of human processing instead of automation.
	
This work focuses on detecting OOD samples in the context of reinforcement learning policies. Several approaches have come up in the recent years to identify OOD data. But, most of them focus on detecting OOD samples for image classification problems. This is due to the fact that the other domains suffer from the availability of OOD or adversarial examples. This work aims to extend the detection of OOD data to deep reinforcement learning and provide a benchmark for future researchers to test their methods.

Our contributions are a benchmark for OOD detection in reinforcement learning, by creating OOD environments that have corrupted observed states or modified physical parameters, which enable the evaluation of OOD detection methods in reinforcement learning. We provide initial results using Dropout, DropConnect, and ensembles, finding that ensembles work best for this task.
    
\section{Related Work}
To the best of our knowledge, there is currently no benchmark available for evaluating out-of-distribution detection in reinforcement learning. However, there has been a surge in interest in this field in the recent years. In this section, we discuss the prior work that has been done for out-of-distribution detection in general. We also discuss the research that has been done for out-of-distribution detection in a reinforcement learning setting.

Various available methods for out-of-distribution or anomaly detection can be categorized based on the availability of anomaly labels into supervised, unsupervised and semi-supervised techniques \cite{bulusu2020} . The supervised methods used uncertainty measures based on the gradient of the negative log-likelihood \cite{Philipp2018}, Mahalanobis distance from different layers \cite{lee2018}, Long short-term memory (LSTM) based binary detectors\cite{Fabio2019}. Some of best semi-supervised methods used Likelihood ratio \cite{ren2019}, Probably Approximately Correct (PAC) based algorithm \cite{Si2018} and a two-head Convolutional Neural Network (CNN) \cite{yu2019} for anomaly detection.

Unsupervised techniques include using predicted softmax probability \cite{Dan2016}, Temperature scaling \cite{Shiyu2017}, and Generative Adversarial Network (GAN) based architecture \cite{Lawson2017}. Overall, supervised methods tend to perform better than the other methods as ground-truths are available. However, having the examples and labels for a full spectrum of anomalies may not be possible in all the cases and this might result in overfitting. Unsupervised methods are flexible and can be applied to a variety of domains as they don't rely on the labels and anomalous data. However, they are highly sensitive to noise. Semi-supervised methods have the flexibility of unlabelled data along with the accuracy from the labelled data. However, they tend to overfit in unseen anomalous situations. One of the challenges in anomaly detection in deep learning is to define the boundary between normal and anomalous examples with complex feature spaces.

Uncertainty estimation provides good results for Independent and Identically Distributed (IID) samples. However, most of these methods tend to fail when there is even a mild change in the dataset distribution. \cite{ovadia2019trust} focuses on understanding the quality of uncertainty estimates in the case of distributional shift along with IID setting. A set of probabilistic deep learning methods like Maximum softmax probability, Monte-Carlo Dropout \cite{gal2016dropout}, Ensembles \cite{lakshminarayanan2016simple}, Temperature Scaling, Stochastic Variational Bayesian Inference (SVI) \cite{wen2018flipout} were evaluated on images, text and MNIST data. In addition to the classification accuracy, metrics like Brier score \cite{Brier1950}, Negative Log-Likelihood and Expected Calibration Error (ECE) are calculated. For MNIST, the accuracy of all the models degrade as the shift in the data increases. The Brier score differentiates the evaluated methods more clearly. All methods have a better Brier score than the state-of-the-art temperature scaling method. Even though SVI achieves the worst accuracy, it outperforms all the other methods when the data shift is significant. Most of the methods show high confidence in their predictions on entirely 
OOD data. CIFAR-10 \cite{alex2012} and ImageNet \cite{imagenet} datasets were used to study the predictive uncertainty on image data. Ensembles had the best performance across most of the metrics. The performance of all the methods follows the same order in both the image datasets. However, the order is not the same as SVI performs worse than vanilla method on the shifted datasets. The 20newsgroups \cite{lang1995} dataset is used to evaluate the predictive uncertainty on text data. Similar to the performance on image data, ensembles outperform all the other methods in terms of accuracy and uncertainty estimation. The uncertainty does not change significantly with temperature scaling even for significantly shifted data. On fully OOD data, vanilla method had better performance than dropout and SVI methods. Overall, ensembles outperformed all the other methods in all the tasks with a better trade-off between accuracy and confidence.

While \cite{bulusu2020} and \cite{ovadia2019trust} provide a large-scale comparison of the OOD methods and an extensive benchmark for evaluating uncertainty estimates respectively, all the discussed methods were evaluated on image or text data. Recently, \cite{andreas2018} presented an OOD detection method applicable for reinforcement learning problems. The solution involves modeling the OOD detection problem as a classification problem with two classes i.e. one for in-distribution data and the other for OOD data. The authors propose a framework called UBOOD \cite{andreas2018} for uncertainty-based OOD detection. It is based on the principle that the epistemic uncertainty is lower for in-distribution (observed during training) samples than for the OOD samples. Two environments were developed for evaluating the proposed framework. One of the environments is a simple grid-based world with a discrete state-space and the other has a continuous state-space based on OpenAI's Lunar lander. Three different versions of the proposed framework have been evaluated based on their F1 scores, with each version based on Monte-Carlo Concrete Dropout network, Bootstrap network, Bootstrap-prior network respectively. The UBOOD framework with Bootstrap-prior network performs the best in detecting the OOD samples on both the environments. It was observed that the F1 score is better for the environment which differs the most from the training environment. Similarly, \cite{safe2019} proposes a risk sensitive reinforcement learning approach that can be combined with a RL policy to make it sensitive to novel data. This work specifically focusses on dynamic obstacle avoidance problem in novel scenarios. The agent can simultaneously observe its goal and the position and velocity of the obstacle. The probabilities of collision for each motion are calculated by LSTM networks. A distribution of the predictions are calculated by MC-Dropout and Bootstrapping. Predictions are used to calculate the mean and variance for each motion primitive. The time to reach the goal after every motion primitive is also estimated in parallel using a simple model. At each time step, the motion primitive with the least collision probability is selected and the process is repeated. This model is evaluated against an uncertainty unaware model. The results show that the uncertainty aware model is more robust to novel obstacles. However, the uncertainty values in novel scenarios did not increase significantly.

While \cite{andreas2018} and \cite{safe2019} explore the implementation of out-of-distribution detection in reinforcement learning tasks, the authors are forced to create their own environments for experiments. This is due to the lack of available benchmark for out-of-distribution or anomaly detection in reinforcement learning. This highlights the need and importance of benchmark tasks for pushing the research on OOD detection in RL even further.

\section{Out-of-Distribution Detection and Uncertainty}

Deep learning is being used to solve complex problems across a variety of domains including autonomous vehicles, industrial automation, health care and surveillance. It has shown to perform as good as or even better than humans in some of these tasks. However, most of the learning methods assume the test data to be from the same distribution as of the training data. This assumption does not hold true in many real world applications. The samples that deviate from the underlying distribution of the training data are referred to as out-of-distribution (OOD) samples or anomalies. Deep learning methods in general are known to be sensitive to OOD data and lead to incorrect results.

An example specific to reinforcement learning is the autonomous control of industrial robots. These robots are typically deployed in a human-robot collaborative environment. In the absence of an OOD detection mechanism, any new work setting which has not been seen during training can make the robot to take actions that could be fatal to the human and other resources that are in its vicinity. This makes out-of-distribution detection extremely important for the safety of the humans and the environment in which the models are deployed.     

Out-of-distribution detection corresponds to the task of identifying samples or observations where the model is uncertain about its output. Different methods used to estimate uncertainty in this work are Monte Carlo Dropout \cite{gal2016dropout}, Monte Carlo DropConnect \cite{mobiny2021dropconnect} and Ensembles \cite{lakshminarayanan2016simple}. Dropout is a regularization technique in neural networks. During training, some of the activations are randomly dropped out. This has proven to be a simple yet effective method to avoid overfitting. Monte Carlo Dropout is the method of enabling dropout at inference which has proven to be an approximation of the predictive posterior distribution.  If the same input is applied to the network multiple times, an empirical distribution can be estimated and the parameters like mean and variance can be obtained. This variance serves as a measure of the model's uncertainty. The variance is expected to be low in the input areas where there was enough training data and can be high where there was no or little training data. 

DropConnect \cite{mobiny2021dropconnect} is a variation of Dropout where the weights are dropped out instead of the activations of a layer. It has also proven to produce an approximation of the predictive posterior distribution. While implementing dropout is simple, implementation of DropConnect requires new layers that use DropConnect layers internally.  MC DropConnect is sometimes seen to perform better than MC Dropout in both learning the task and the uncertainty quantification. Ensemble corresponds to training multiple instances of the same model but randomly drawn initial weights and then combining the predictions. They have better prediction performance than any single member model. They have also shown to exhibit excellent uncertainty estimation properties.For classification tasks, the entropy can be used as a measure of uncertainty. For regression tasks, the standard deviation of the output can be used.

\begin{figure}[t]
    \centering
    \begin{subfigure}{0.19\textwidth}
        \centering
        \includegraphics[width=0.9\linewidth]{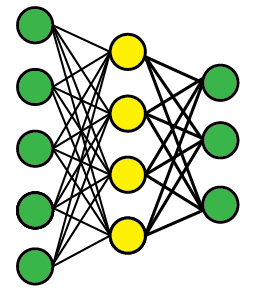}
        \caption{Standard network}
    \end{subfigure}\hfill
    \begin{subfigure}{0.39\textwidth}
        \centering
        \includegraphics[width=0.9\linewidth]{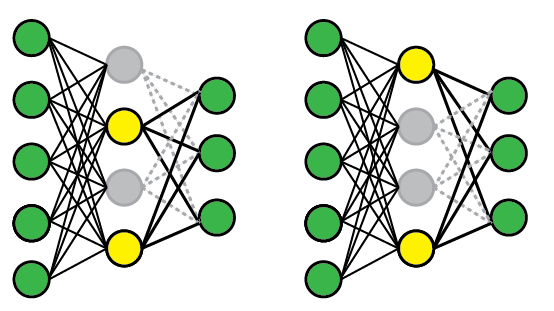}
        \caption{Network after applying dropout}
    \end{subfigure}
    \begin{subfigure}{0.39\textwidth}
        \centering
        \includegraphics[width=0.9\linewidth]{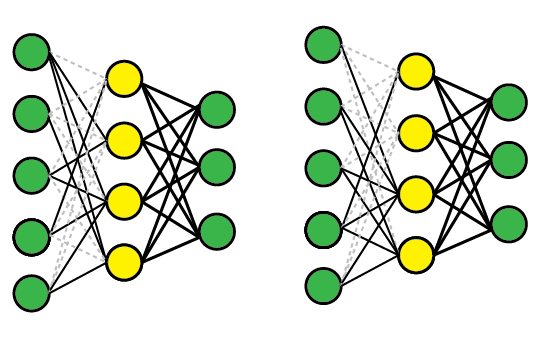}
        \caption{Network after applying dropconnect}
    \end{subfigure}
    \caption{Figure showing the effect of using dropout layers in the network \cite{dropoutimage}}
\end{figure}

\begin{figure}[t]
\centering
\includegraphics[width=0.5\textwidth]{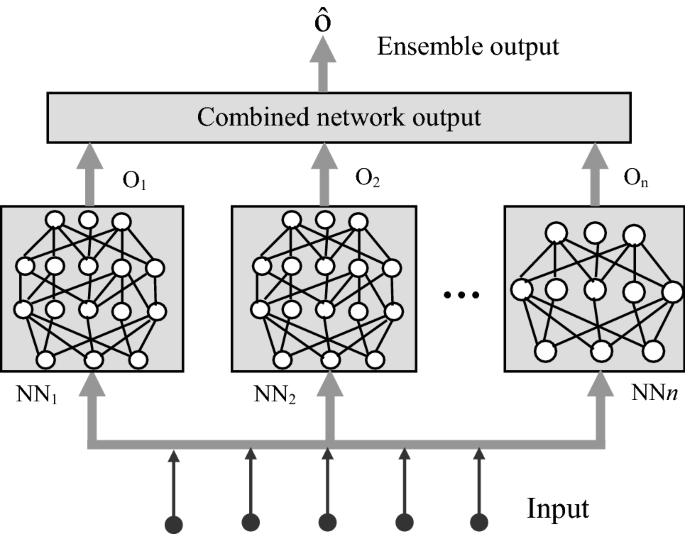}
\caption{Training an Ensemble network \cite{ensembleimage}}
\end{figure}

\section{Out of Distribution Detection in Deep RL Benchmark}

To generate custom versions of the cartpole environment, physical variables like gravity, mass of the cart, mass of the pole, length of the pole and the magnitude of the force to be applied to the cart are assigned new values. A grid of values are chosen in multiples of the default values for each parameter. For example, the force magnitude has a default value of 10. The custom versions have values ranging from 1.0 to 100.0 as a multiple (x/10, x/9, x/8,..., x/2, 2x, 3x,...,10x) of the default value. Custom versions of the pendulum environment are generated in the same way as the cartpole environment by assigning new values of physical variables like gravity, mass of the pole, length of the pole along with the speed and torque to be applied. Unlike the physical environments like cartpole and pendulum, custom versions of the pong environment are generated by corrupting the observations (images). For this, we have used the imagecorruptions \cite{imagecorruptions} package, that supports various corruption types including gaussian noise, impulse noise, motion blur, and pixelate. We can also adjust the severity of the corruption using this package. 

We present all the parameters values for the custom environments that should make them out of distribution. Table \ref{all-custom-envs} lists the physical variables, their default values and the new values that are assigned to generate custom versions of for all environments.

\begin{table}[t]
    \centering
    \begin{tabular}{@{}llp{7cm}@{}}
        \toprule
        Parameter          & Default value & OOD Environment - Parameter Values \\
        \midrule
        \multicolumn{2}{l}{\textbf{Cartpole - Physical Parameters}} & \textbf{OOD-Cartpole}\\
        \midrule
        Gravity            & 9.8           & 0.98, 1.09, 1.23, 1.4, 1.63, 1.96, 2.45, 3.27, 4.9, 19.6, 29.4, 39.2, 49.0, 58.8, 68.6, 78.4, 88.2, 98.0\\
        Mass of the cart   & 1.0           & 0.1, 0.1111, 0.125, 0.1429, 0.1667, 0.2, 0.25, 0.3333, 0.5, 2.0, 3.0, 4.0, 5.0, 6.0, 7.0, 8.0, 9.0, 10.0\\
        Length of the pole & 0.5           & 0.05, 0.0556, 0.0625, 0.0714, 0.0833, 0.1, 0.125, 0.1667, 0.25, 1.0, 1.5, 2.0, 2.5, 3.0, 3.5, 4.0, 4.5, 5.0\\
        Mass of the pole   & 0.1           & 0.01, 0.0111, 0.0125, 0.0143, 0.0167, 0.02, 0.025, 0.0333, 0.05, 0.2, 0.3, 0.4, 0.5, 0.6, 0.7, 0.8, 0.9, 1.0\\
        Force magnitude    & 10.0          & 1.0, 1.1111, 1.25, 1.4286, 1.6667, 2.0, 2.5, 3.3333, 5.0, 20.0, 30.0, 40.0, 50.0, 60.0, 70.0, 80.0, 90.0, 100.0 \\
        \midrule
        \multicolumn{2}{l}{\textbf{Pendulum - Physical Parameters}} & \textbf{OOD-Pendulum}\\
        \midrule
        Gravity            & 10.0          & 0.5, 1.0, 2.0, 5.0, 20.0, 50.0, 100.0, 200.0 \\
        Mass of the pole   & 1.0           & 0.05, 0.1, 0.2, 0.5, 2.0, 5.0, 10.0, 20.0\\
        Length of the pole & 1.0           & 0.05, 0.1, 0.2, 0.5, 2.0, 5.0, 10.0, 20.0\\
        Max speed          & 8.0           & 0.4, 0.8, 1.6, 4.0, 16.0, 40.0, 80.0, 160.0\\
        Max torque         & 2.0           & 0.1, 0.2, 0.4, 1.0, 4.0, 10.0, 20.0, 40.0\\
        \midrule
        \multicolumn{2}{l}{\textbf{Pong - State Observation Corruptions}} & \textbf{OOD-Pong}\\
        \midrule
        Gaussian noise $\sigma$ 		& 0.0 & 0.08, 0.12, 0.18, 0.26, 0.38 \\
        Impulse noise $p$  				& 0.0 & 0.03, 0.06, 0.09, 0.17, 0.27 \\
        Motion blur $\rho, \sigma$  	&(0, 0) & (10, 3), (15, 5), (15, 8), (15, 12), (20, 15) \\
        Pixelate downscale factor $f$	& 1.0 & 0.6, 0.5, 0.4, 0.3, 0.25 \\
        \bottomrule
    \end{tabular}
    \vspace*{0.2em}
    \caption{A list of physical and image corruption parameters for all custom environments along with their default values (used for training) and new values used for out of distribution detection evaluation.}
    \label{all-custom-envs}
\end{table}

Unlike the above mentioned physics based environments like cartpole and pendulum, custom versions of the pong environment are generated by corrupting the observations (images). Table \ref{all-custom-envs} lists the corruption types and their parameters used to generate custom versions of the pong environment. The appendix contains details of the image/state corruptions and their formulations.

\section{Experimental Setup}
 In this section, various tasks that were used to evaluate the performance of out-of-distribution detection methods are described. As real world applications of reinforcement learning are in both visual and non-visual based environments, a combination of tasks that cover these aspects are chosen. These tasks include cartpole, pendulum and pong. Cartpole is a non-visual physics based environment and has a discrete action space. Pendulum is also a non-visual physics based environment but with a continuous action space. Pong is a visual based environment with a discrete action space. These environments have been taken from OpenAI Gym suite \cite{gym} which is a framework for the development and comparison of reinforcement learning algorithms.
 
 \begin{figure}[t]
    \centering
    \begin{subfigure}{0.2\textwidth}
        \centering
        \includegraphics[height = 0.13\textheight]{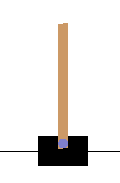}
        \caption{Cartpole}
    \end{subfigure}\hfill
    \begin{subfigure}{0.35\textwidth}
        \centering
        \includegraphics[height = 0.13\textheight]{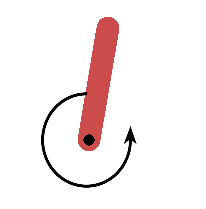}
        \caption{Pendulum}
    \end{subfigure}
    \begin{subfigure}{0.35\textwidth}
        \centering
        \includegraphics[height = 0.13\textheight]{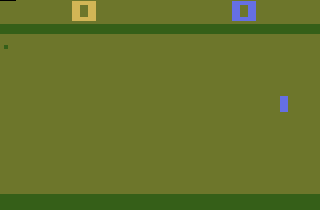}
        \caption{Pong}
    \end{subfigure}\hfill
    \caption{Figures showing the different environments/tasks used.}
\end{figure}
 
The performance of the different out-of-distribution detection methods is evaluated in the following way.
\begin{enumerate}
    \item A task is trained using a deep RL algorithm.
    \item The trained model is evaluated on multiple custom versions of the environment.
    \item The custom environment versions where the trained model fails are identified.
    \item Different out-of-distribution detection methods are evaluated on these custom environments.
    \item The above steps are repeated for different tasks.
\end{enumerate}
    
\begin{table}[t]
\centering
\begin{tabular}{@{}lll@{}}
\toprule
Task     & Algorithm & OOD detection method                 \\ \midrule
Cartpole & DQN       & MC Dropout, MC DropConnect, Ensemble \\
Pendulum & DDPG      & MC Dropout, MC DropConnect, Ensemble \\
Pong     & DQN       & MC DropConnect, Ensemble             \\ \bottomrule
\end{tabular}
\vspace*{0.2em}
\caption{Combinations of tasks, RL algorithms and OOD detection methods evaluated.}
\label{ood-combinations}
\end{table}

\section{Experimental Results and Analysis}

In this section, we analyze the OOD detection performance of various methods on cartpole, pendulum and the pong environments.
	
Table \ref{cartpole-best-auc} lists the different OOD methods along with their best AUC scores on the specific custom versions of the cartpole environment. MC Dropout achieves its best AUC score of 0.78 on the custom version having the force parameter of 2.5. On the other hand, MC DropConnect achieves its best AUC score of 0.992 on the custom version having the gravity of 78.4. However, the best overall AUC score of 0.993 on the custom cartpole environments is achieved by the ensemble method on the version with the length of the pole as 2. Apart from the best AUC scores, the lowest standard deviation of the AUC scores across trials is also achieved by the ensemble method followed by the MC Dropout and then the MC DropConnect which has the largest standard deviation values for the custom cartpole environments. This shows that the Ensemble method is the best performing OOD detection method for the cartpole environment. 

\begin{table}[t]
\centering
\begin{tabular}{@{}llc@{}}
\toprule
\multicolumn{1}{c}{OOD detection method} & \multicolumn{1}{c}{Parameters} & Best AUC score                \\
\midrule
MC Dropout                               & Force: 2.5                               & \textbf{0.780}                         \\
MC Dropout                               & Gravity: 49                              & 0.759                         \\
MC Dropout                               & Mass of the cart: 3                      & 0.719                         \\
\midrule
MC DropConnect                           & Force: 1                                 & 0.921 \\
MC DropConnect                           & Gravity: 78.4                            & \textbf{0.992}                        \\
MC DropConnect                           & Length of the pole: 2                    & 0.987                         \\
MC DropConnect                           & Mass of the cart: 9                      & 0.887                         \\
\midrule
Ensemble                                 & Gravity: 98                              & 0.833                         \\
Ensemble                                 & Length of the pole: 2           & \textbf{0.993}                         \\
\bottomrule
\end{tabular}
\vspace*{0.2em}
\caption{Best AUC scores achieved by MC Dropout, MC DropConnect and ensemble method on variations of OOD-Cartpole. The overall best AUC score of each OOD method is highlighted.}
\label{cartpole-best-auc}
\end{table}

\begin{figure}[t]
\centering
\includegraphics[width=0.95\textwidth]{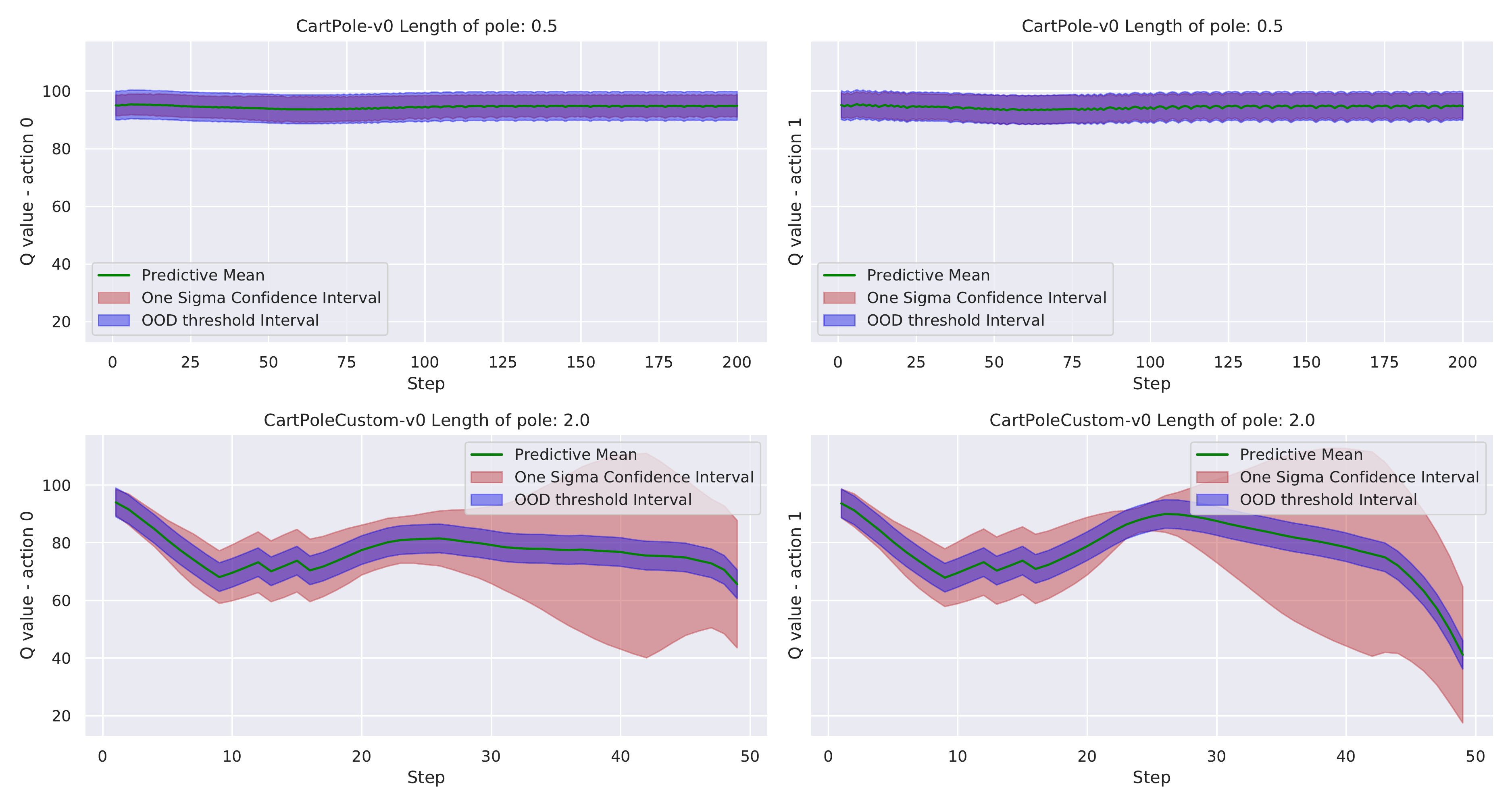}
\caption{Comparison of Ensemble mean and standard deviation of Q values produced by DQN along with the threshold for OOD detection for Cartpole and OOD-Cartpole length of pole set at two.}
\label{cartpole-dqn-ensemble-std-length-2}
\end{figure}

Figure \ref{cartpole-dqn-ensemble-std-length-2} show the mean and standard deviation of the Q values during one episode for both the original environment and the custom version with a length of 2. As seen from the figure, the standard deviation is almost always less than the OOD threshold for the original environment over the course of the entire episode. However, for the custom environment, the standard deviation is more than the OOD threshold right from the initial steps of the episode. This behavior is seen for both the actions. This also highlights the good OOD detection performance of the ensemble method.
	
Table \ref{pendulum-best-auc} lists the different OOD methods along with their best AUC scores on the specific custom versions of the pendulum environment. MC Dropout achieves its best AUC score of 0.727 on the custom version having the gravity of 50. On the other hand, MC DropConnect achieves its best AUC score of 0.726 on the custom version having the length of the pole of 5. Similarly, the ensemble method achieves its best overall AUC score of 0.619 on the custom pendulum environment having the length of the pole as 0.1. Apart from the best AUC scores, the lowest standard deviation of the AUC scores across trials is achieved by the ensemble method followed by the MC Dropout and then the MC DropConnect which has the largest standard deviation values for the custom pendulum environments. This shows that there is a trade off between the best possible performance and consistency for the OOD detection methods. Overall, the MC Dropout method can be considered the best performing OOD detection method for the pendulum environment with an acceptable level of reproducibility in performance. 

\begin{table}[t]
\centering
\begin{tabular}{@{}llc@{}}
\toprule
\multicolumn{1}{c}{OOD detection method} & \multicolumn{1}{c}{Custom configuration} & Best AUC score                \\ \midrule
MC Dropout                               & Gravity: 50                              & \textbf{0.727}                       \\
MC Dropout                               & Length of the pole: 5                    & 0.656                         \\
MC Dropout                               & Mass of the cart: 5                      & 0.726 \\
MC DropConnect                           & Gravity: 50                              & 0.602                         \\
MC DropConnect                           & Length of the pole: 5                    & \textbf{0.726 }                        \\
MC DropConnect                           & Mass of the cart: 5                      & 0.715                         \\
Ensemble                                 & Length of the pole: 0.1                  & \textbf{0.619}                         \\
Ensemble                                 & Mass of the cart: 0.05          & 0.596                         \\ \bottomrule
\end{tabular}
\vspace*{0.2em}
\caption{Best AUC scores achieved by MC Dropout, MC DropConnect and ensemble method on variations of OOD-Pendulum. The overall best AUC score of each OOD method is highlighted.}
\label{pendulum-best-auc}
\end{table}

\begin{figure}[t]
\centering
\includegraphics[width=0.7\textwidth]{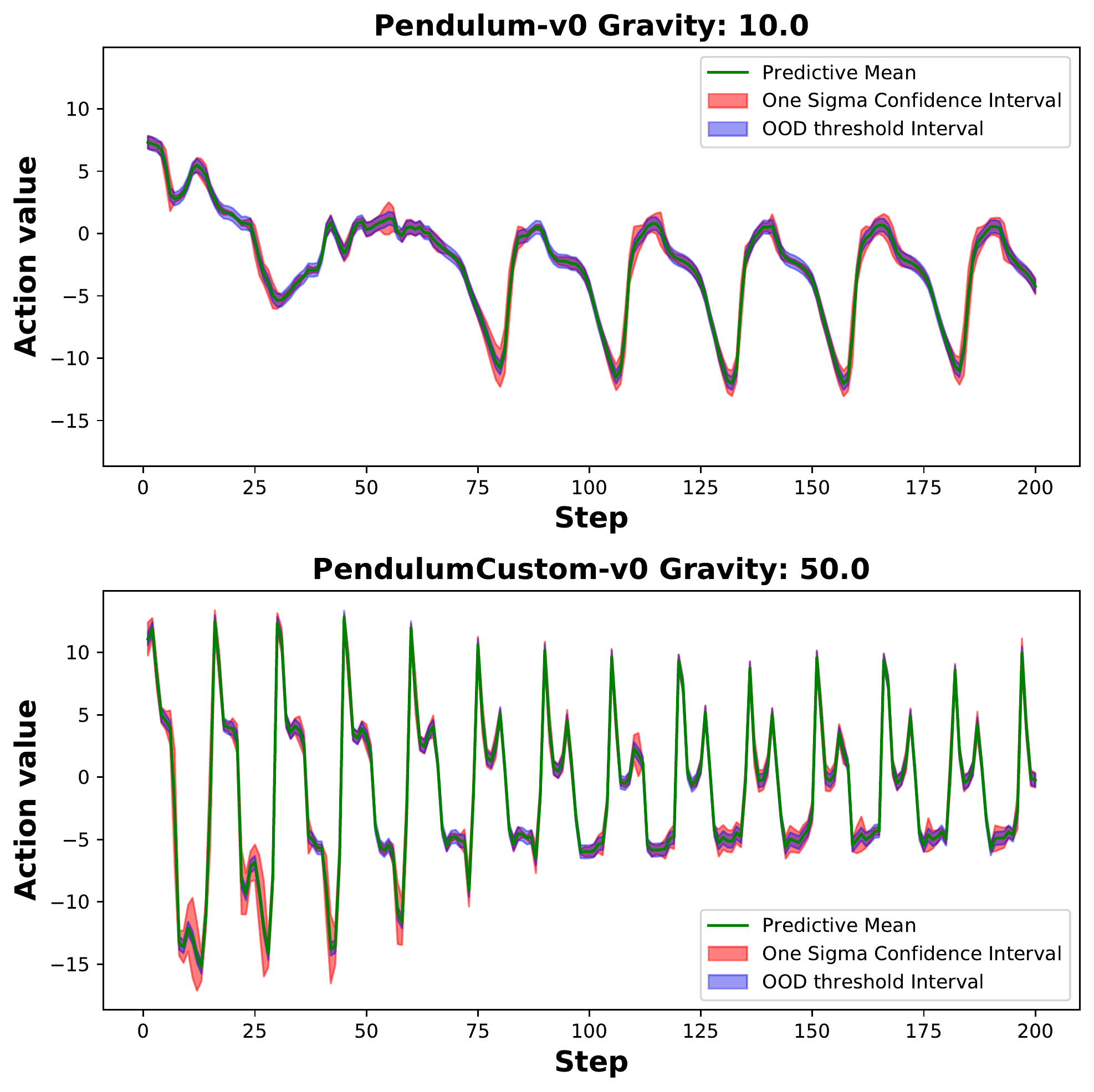}
\caption{Comparison of action values obtained by MC Dropout on DDPG for original pendulum environment with gravity of 10 and the custom version with gravity of 50. The plot shows the mean of the action value obtained using the five predictors along with the standard deviation and the best threshold interval computed to distinguish the ID and OOD observations.}
\label{pendulum-ddpg-dropout-std}
\end{figure}

Figure \ref{pendulum-ddpg-dropout-std} shows the progress of the action values along with their standard deviations for a duration of one episode obtained during the best performing trail of MC Dropout model for the original pendulum environment with gravity of 10 and the custom version with gravity of 50.  It can be seen that the standard deviation values obtained for the original pendulum environment are higher than the OOD threshold value especially in the middle of the episode. Similarly, the standard deviation values obtained for the custom environment are higher than the OOD threshold value in the beginning of the episode and also occasionally in the later part of the episode. This shows that MC Dropout is not very efficient in distinguishing between ID and OOD observations for the pendulum environment.
	
Table \ref{pong-best-auc} lists the different OOD methods along with their best AUC scores on the specific custom versions of the pong environment.  MC DropConnect achieves its best AUC score of 0.828 on the custom pong environment corrupted with Gaussian noise of severity level 5. On the other hand, the ensemble method achieves its best AUC score of 0.91 on the custom version corrupted using motion blur with a severity of 3. When the performance of both the methods is compared based on the corruption type, the ensemble method achieves better AUC scores than MC DropConnect across all corruptions. The ensemble method also has lower standard deviations across trials than MC DropConnect. This shows that the ensemble method has the best OOD detection performance for the pong environment.

\begin{table}[t]
\centering
\begin{tabular}{@{}lcl@{}}
\toprule
\multicolumn{1}{c}{}                                       & \multicolumn{2}{c}{Best AUC score}       \\ \cmidrule(l){2-3} 
\multicolumn{1}{c}{\multirow{-2}{*}{Custom configuration}} & MC DropConnect                & Ensemble \\
\midrule
Gaussian noise: $\sigma = 0.18$                            & 0.699                         & 0.647    \\
Gaussian noise: $\sigma = 0.26$                            & 0.749 						   & 0.735    \\
Gaussian noise: $\sigma = 0.38$                            & \textbf{0.828}                & 0.831    \\
\midrule
Impulse noise: $p = 0.09$                                  & 0.712                         & 0.656    \\
Impulse noise: $p = 0.17$                                  & 0.775                         & 0.769    \\
Impulse noise: $p = 0.27$                                  & 0.822                         & 0.836    \\
\midrule
Motion blur: $\rho = 15, \sigma = 8$                       & 0.707                         & \textbf{0.910}    \\
Motion blur: $\rho = 15, \sigma = 12$                      & 0.682                         & 0.861    \\
Motion blur: $\rho = 20, \sigma = 15$                      & 0.685                         & 0.841    \\
\midrule
Pixelate: $f = 0.4$                                        & 0.606                         & 0.823    \\
Pixelate: $f = 0.3$                                        & 0.565                         & 0.634    \\ 
\bottomrule
\end{tabular}
\vspace*{0.2em}
\caption{Best AUC scores achieved by MC DropConnect and Ensemble methods on variations of OOD-Pong.}
\label{pong-best-auc}
\end{table}

\newpage
\section{Conclusions and Future Work}

In this work, the OOD detection performance of different uncertainty estimation methods i.e. MC Dropout, MC DropConnect and ensemble is compared across a range of control tasks. The tasks included two physics based environments i.e. cartpole, which has a discrete action space and pendulum, which has a continuous action space. The OOD detection methods were also evaluated on pong, which is a visual based environment. The difference in the performance of the trained models between the original environment and  the custom versions highlight the sensitivity of the models to the changes in the environment. The models trained on visual based environment were, in general, more sensitive to changes in the environment than the models trained on physics based environments.  One of the major challenges faced during training the dropout and dropconnect models was to identify the appropriate level of dropout probability that the models can still learn to solve the original versions of the tasks.
	
Ensemble methods achieved the best OOD detection performance on cartpole and pong environments while MC Dropout performed the best on the pendulum environment. The ensemble method also had the lowest variation in its performance over multiple trials across all the environments. The MC DropConnect has a good OOD detection performance across all the environments, however, it is not consistent. This work also highlights the effect of the RL algorithm on the performance of the OOD detection methods.  The overall AUC scores obtained using DQN based models are higher than the ones obtained by DDPG.  Nonetheless, more experiments are needed to confirm this behavior. 
	
Overall, the experiments show that MC Dropout, MC DropConnect and the ensemble method were successful in detecting OOD observations in deep reinforcement learning.  This is especially true in the case of the visual based environments, where the trained models failed on almost all the custom versions of the environment but were able to detect OOD observations to a large extent especially with higher levels of corruption. Future work can be done in developing methods that not only estimate the uncertainty but also learn from the OOD observations to create more robust models.  Researchers are also encouraged to test the OOD detection methods on more complex environments.
	
\bibliography{bibliography}

\appendix
\clearpage


\section{Image/State Observation Corruptions}

\begin{figure}[!htb]
    \centering
    \begin{subfigure}{0.24\textwidth}
        \includegraphics[width=\linewidth]{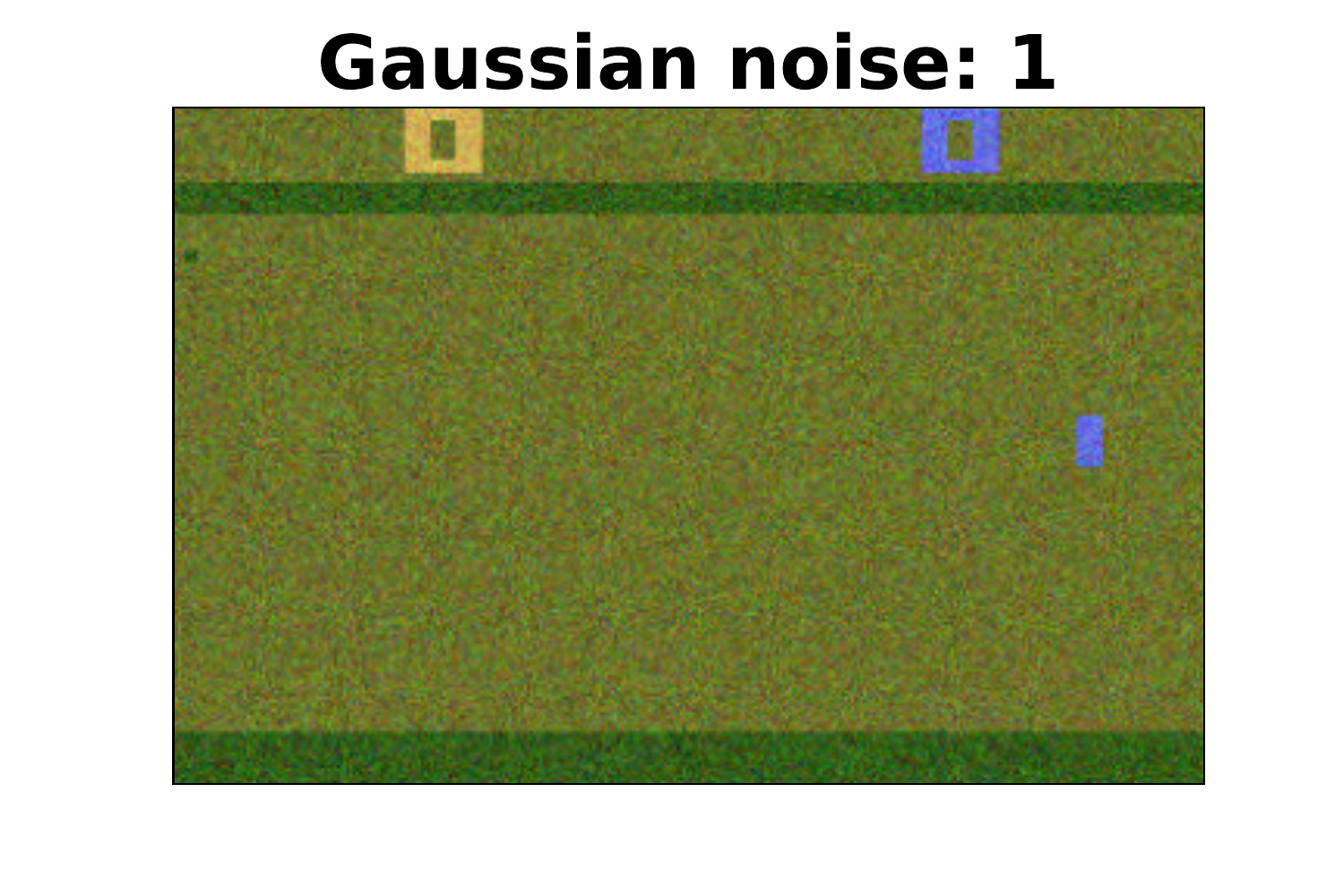}
        \caption{$\sigma = 0.08$}
    \end{subfigure}
    \begin{subfigure}{0.24\textwidth}
        \includegraphics[width=\linewidth]{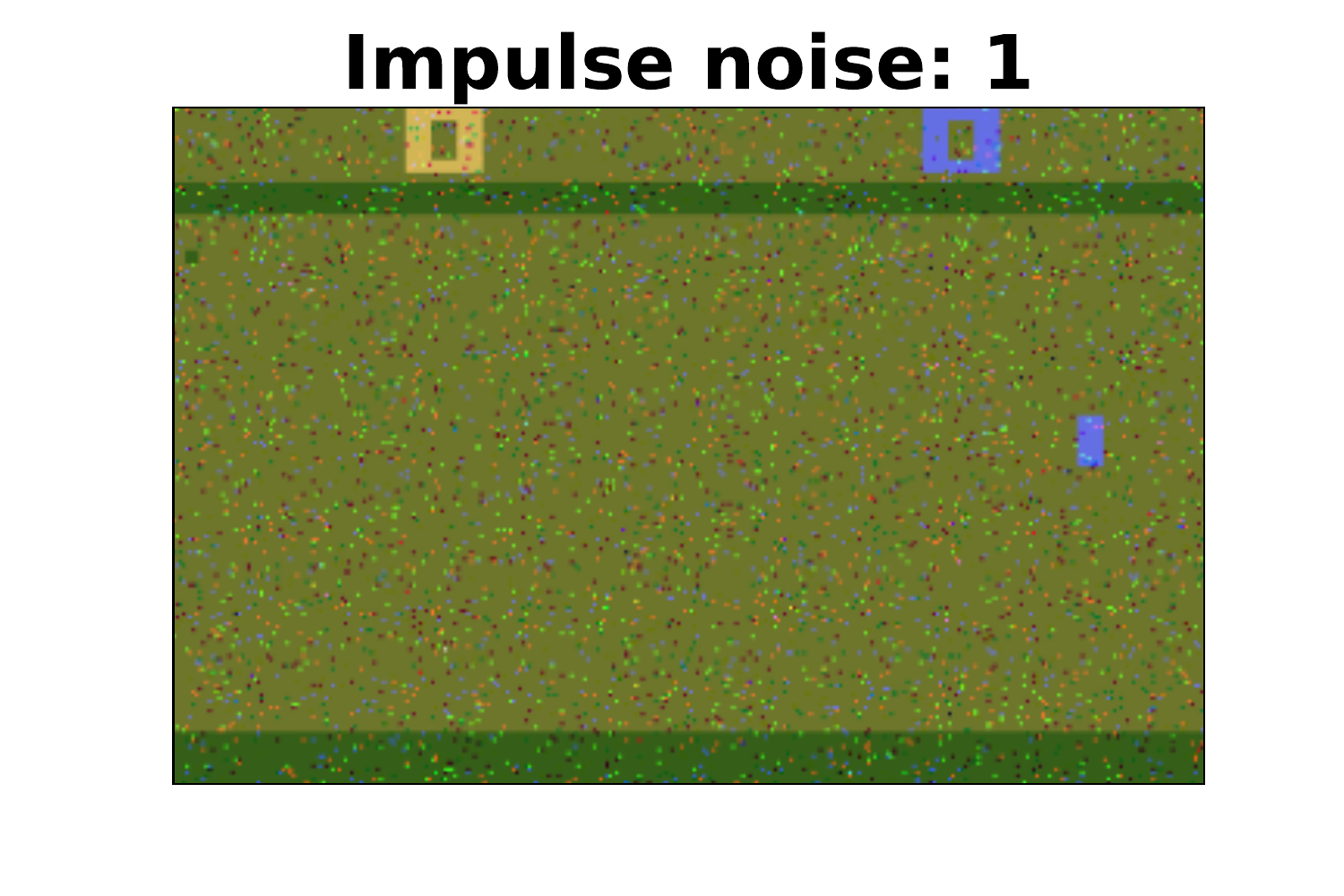}
        \caption{$p = 0.03$}
    \end{subfigure}
    \begin{subfigure}{0.24\textwidth}
        \includegraphics[width=\linewidth]{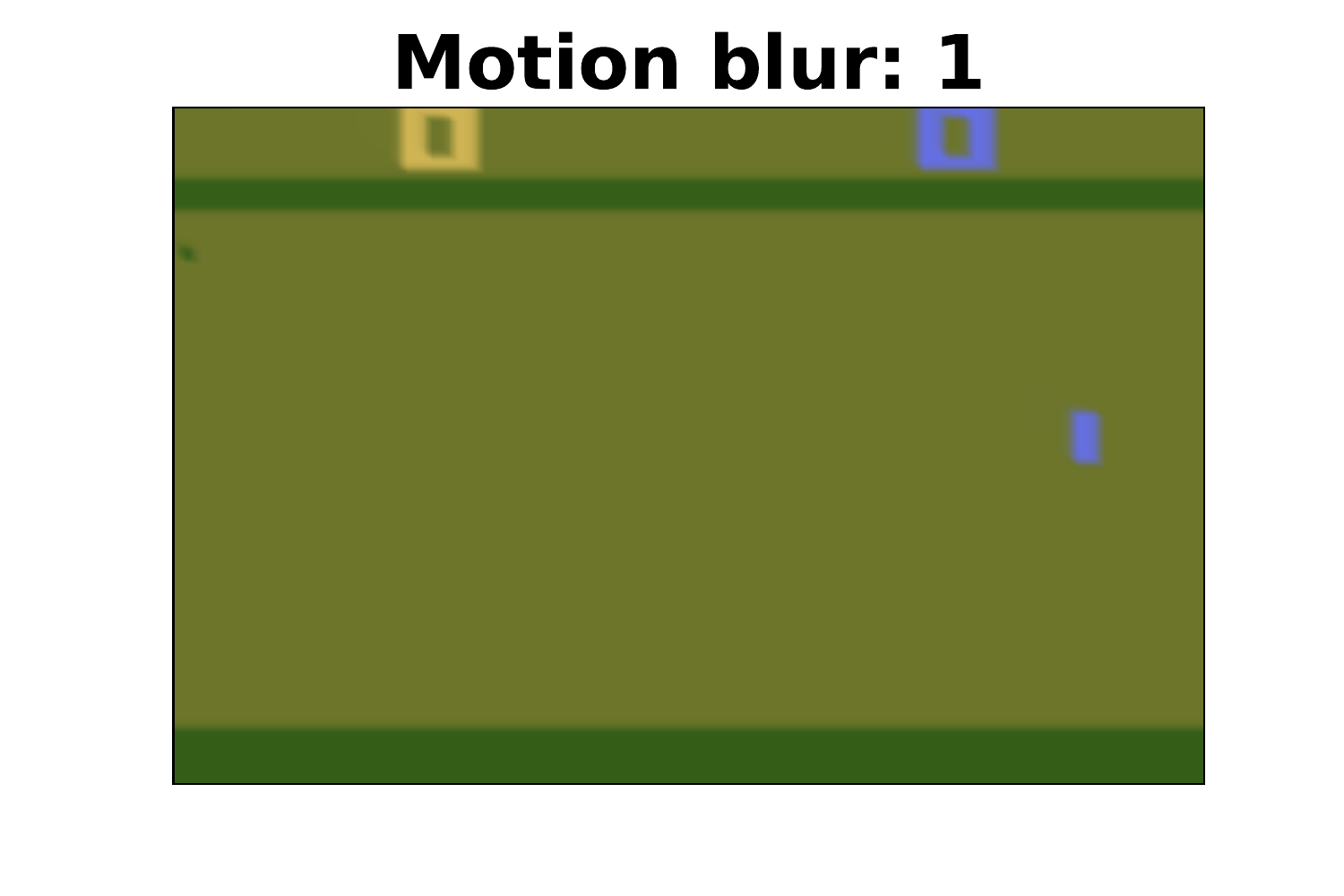}
        \caption{$\rho = 10\sigma = 3$}
    \end{subfigure}
    \begin{subfigure}{0.24\textwidth}
        \includegraphics[width=\linewidth]{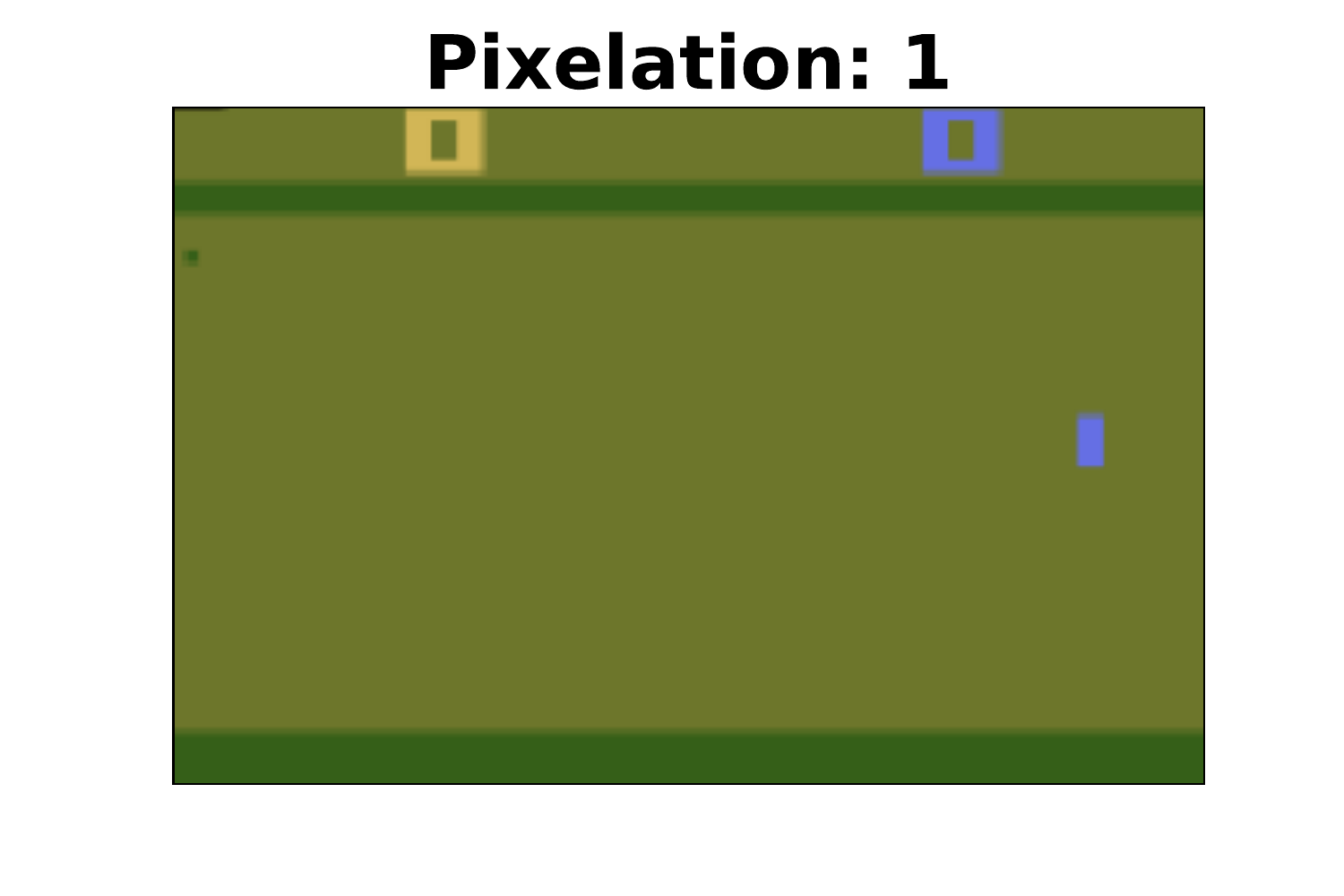}
        \caption{$f = 0.6$}
    \end{subfigure}
    
    \begin{subfigure}{0.24\textwidth}
        \includegraphics[width=\linewidth]{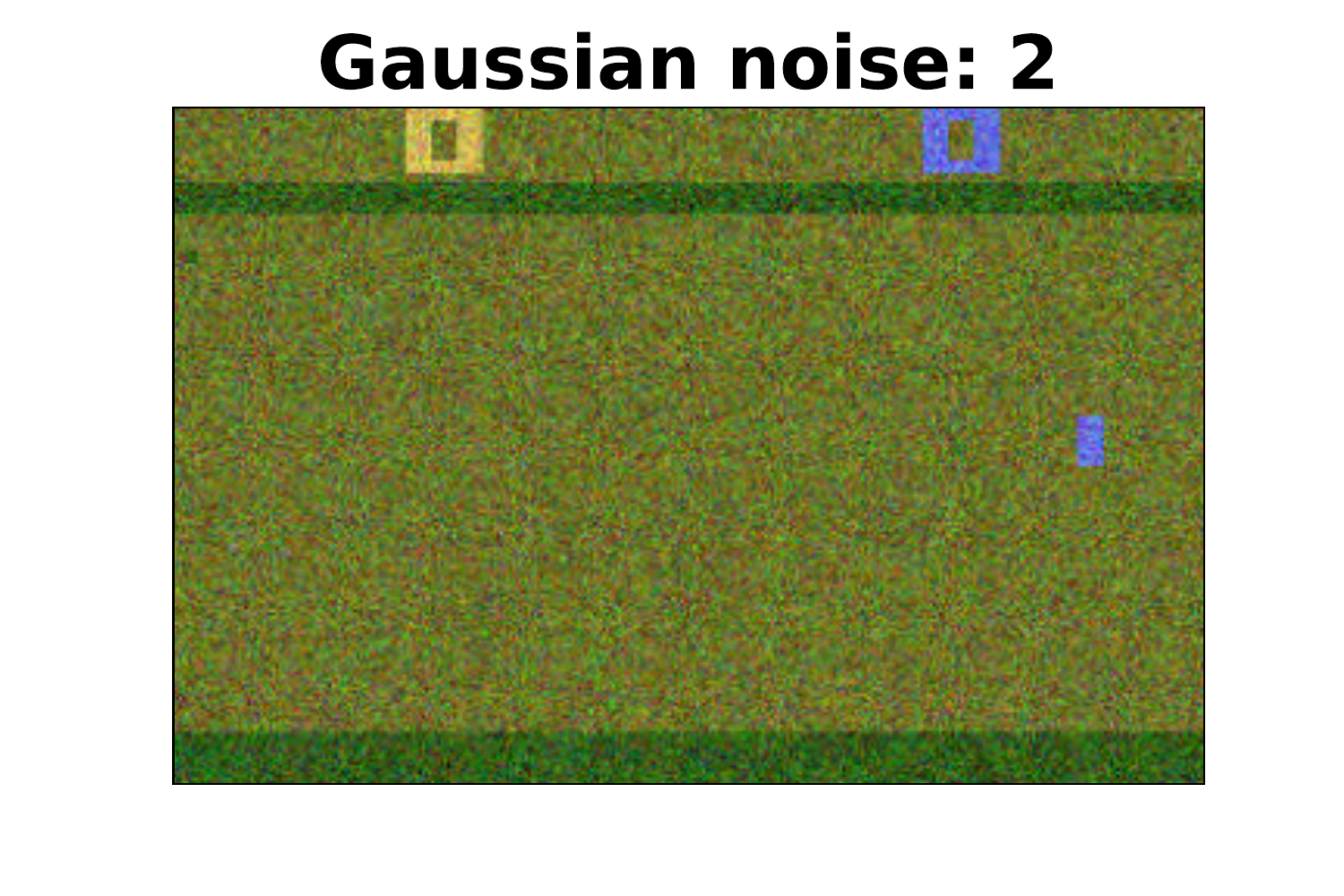}
        \caption{$\sigma = 0.12$}
    \end{subfigure}
    \begin{subfigure}{0.24\textwidth}
        \includegraphics[width=\linewidth]{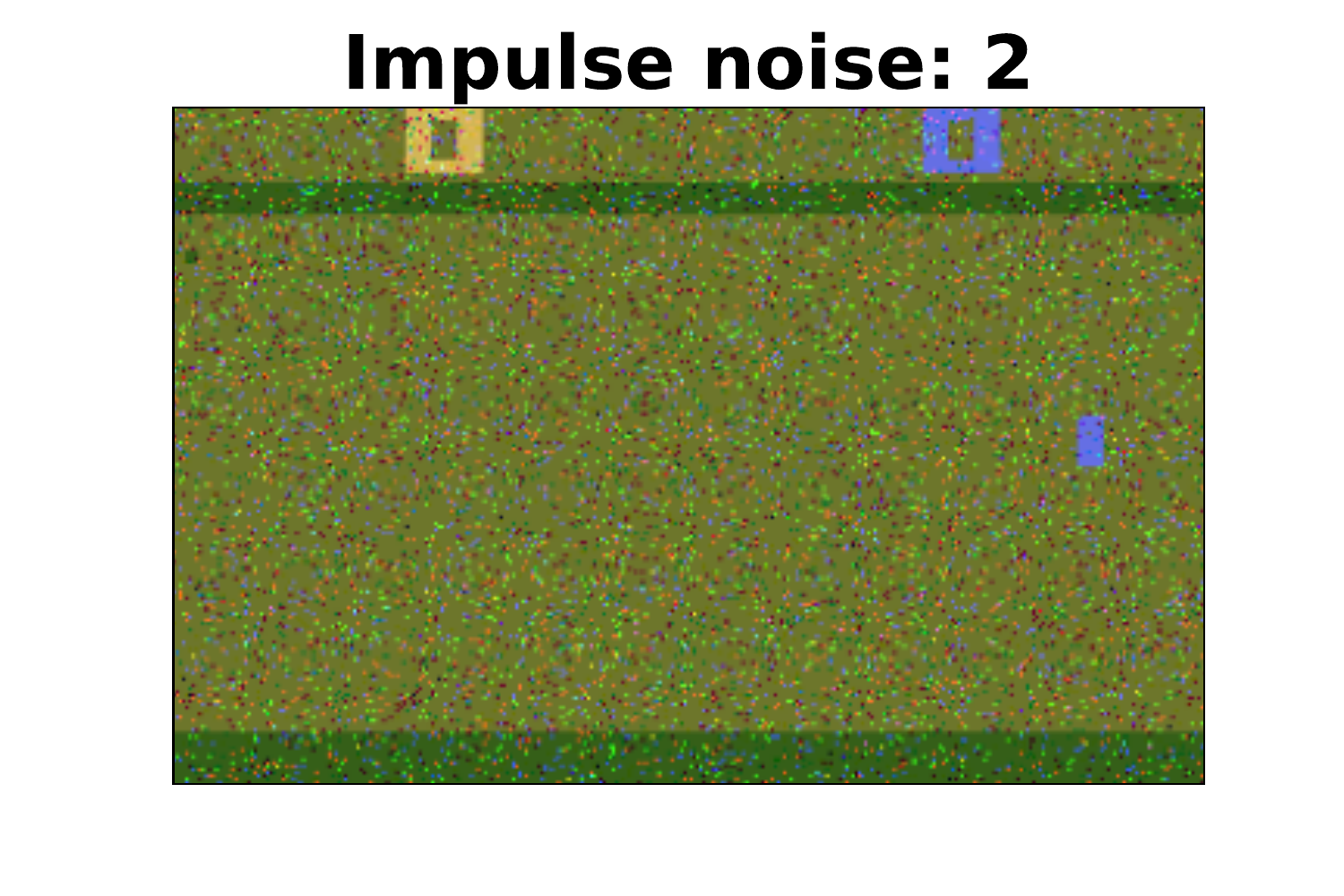}
        \caption{$p = 0.06$}
    \end{subfigure}
    \begin{subfigure}{0.24\textwidth}
        \includegraphics[width=\linewidth]{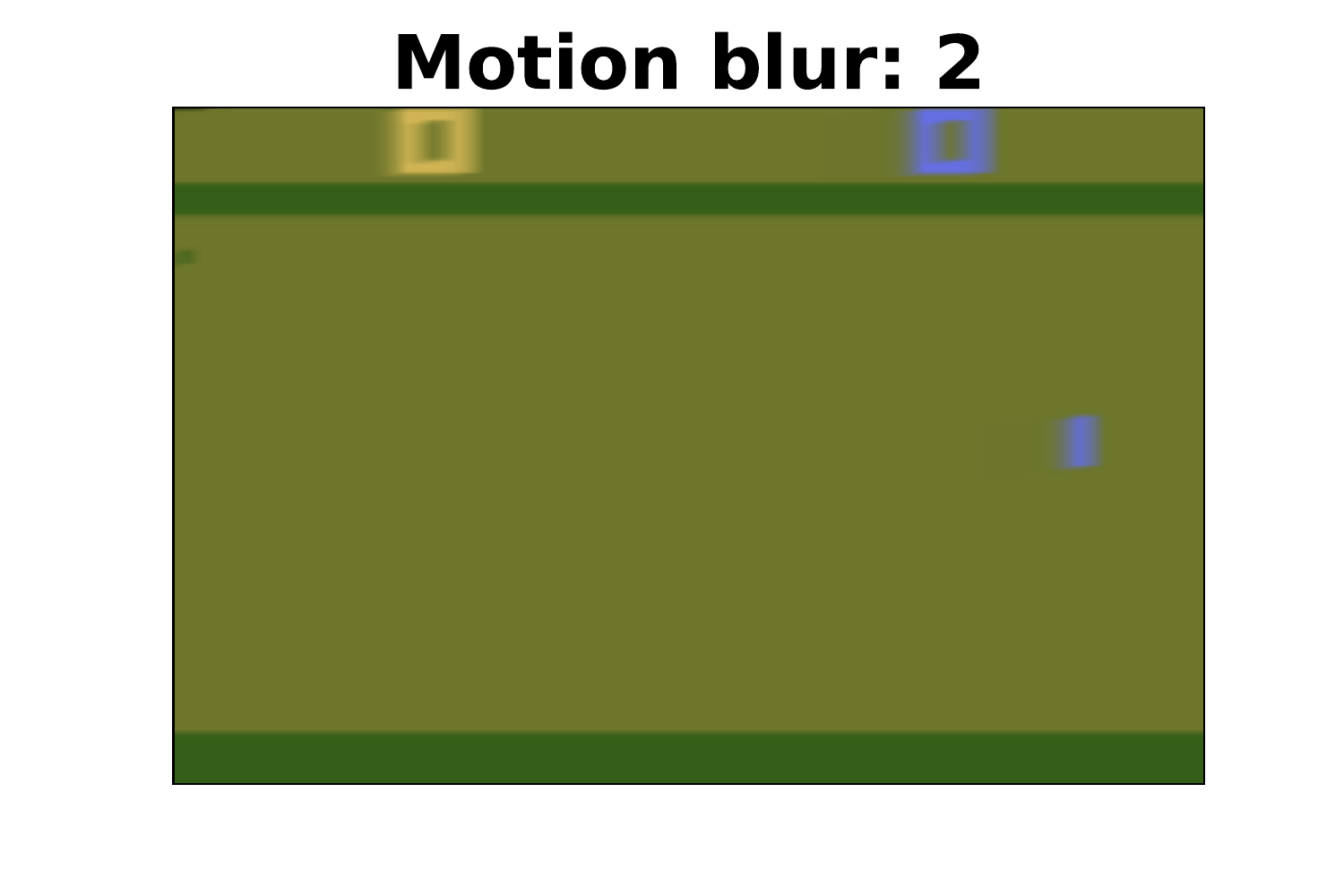}
        \caption{$\rho = 15\sigma = 5$}
    \end{subfigure}
    \begin{subfigure}{0.24\textwidth}
        \includegraphics[width=\linewidth]{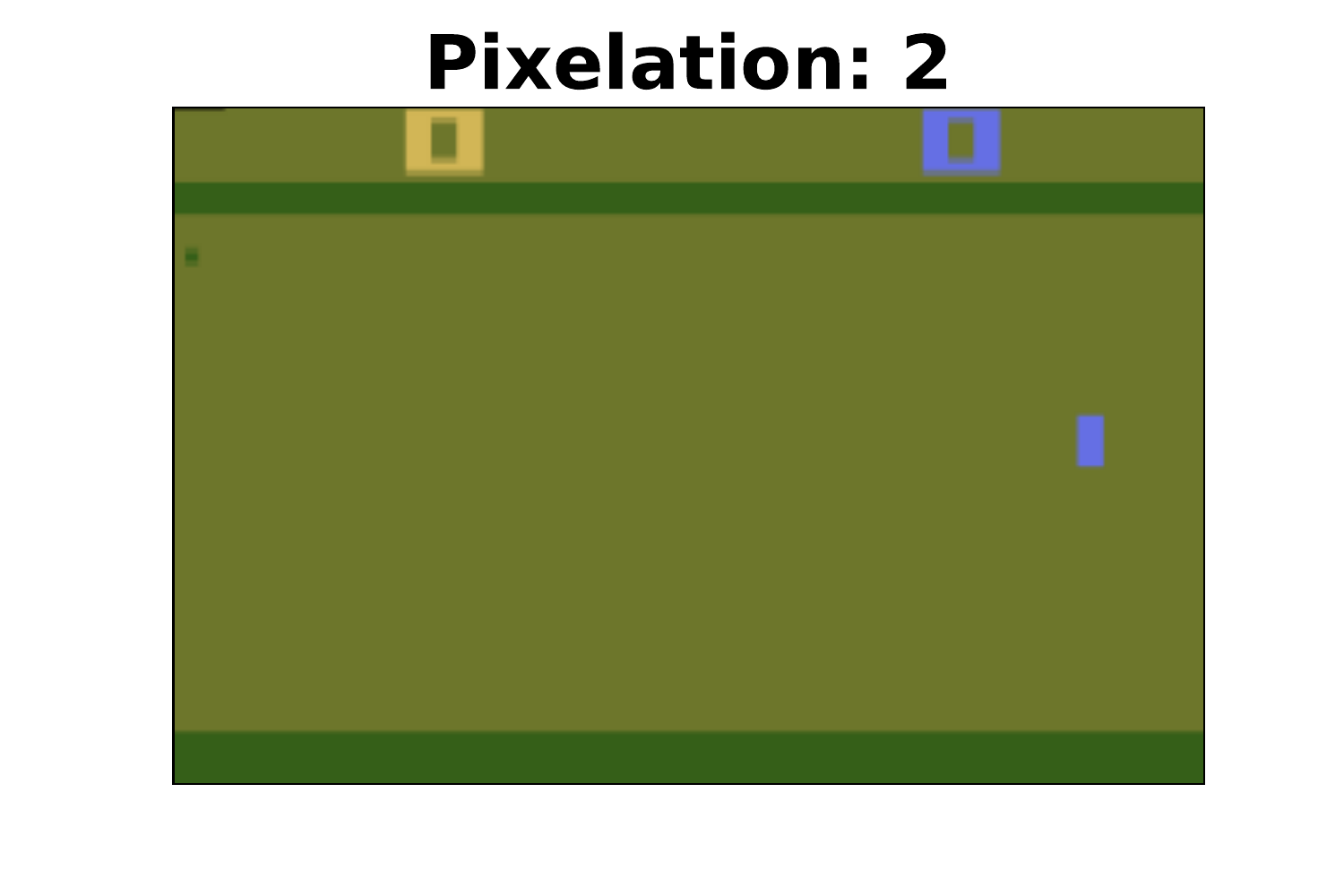}
        \caption{$f = 0.5$}
    \end{subfigure}

    \begin{subfigure}{0.24\textwidth}
        \includegraphics[width=\linewidth]{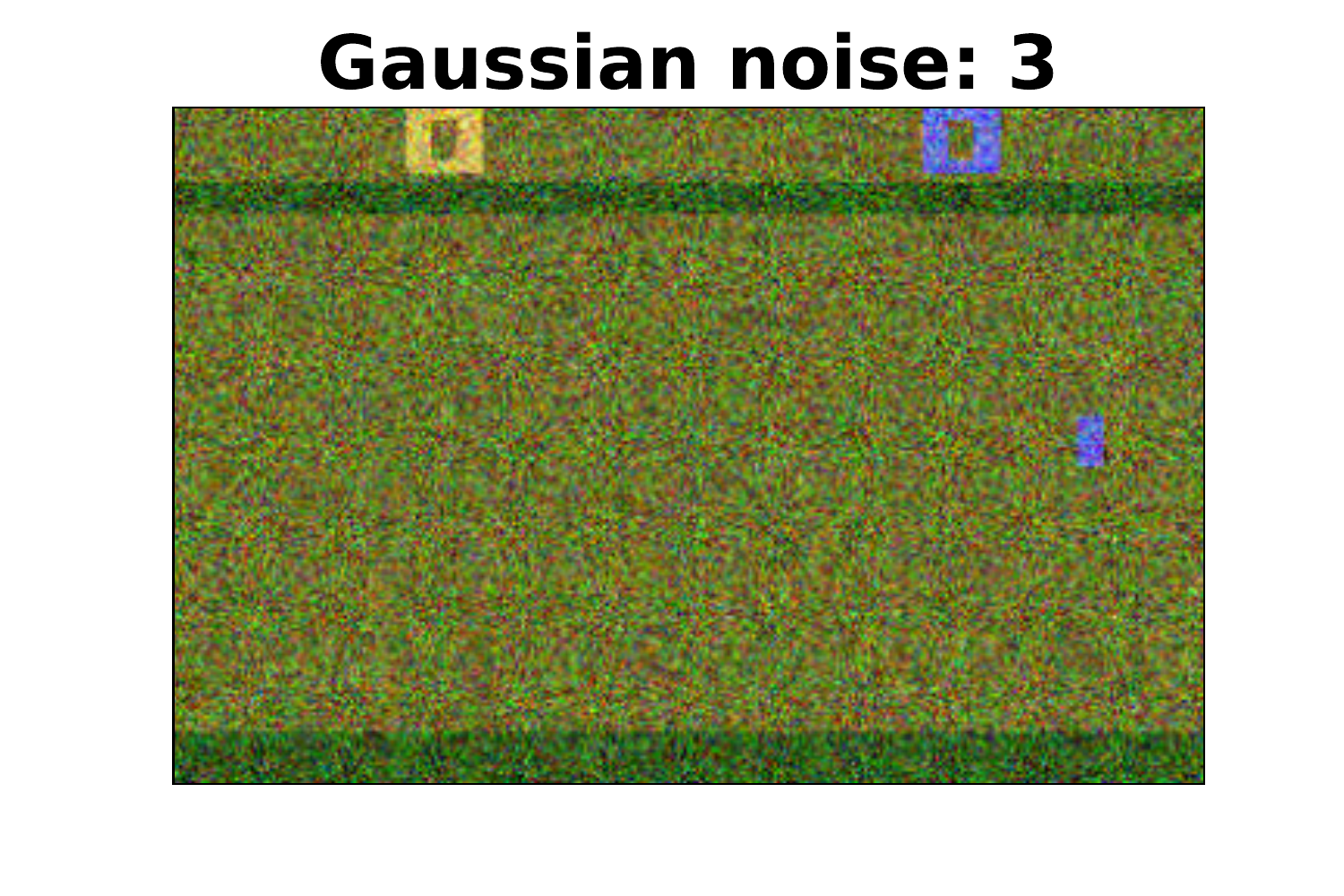}
        \caption{$\sigma = 0.18$}
    \end{subfigure}
    \begin{subfigure}{0.24\textwidth}
        \includegraphics[width=\linewidth]{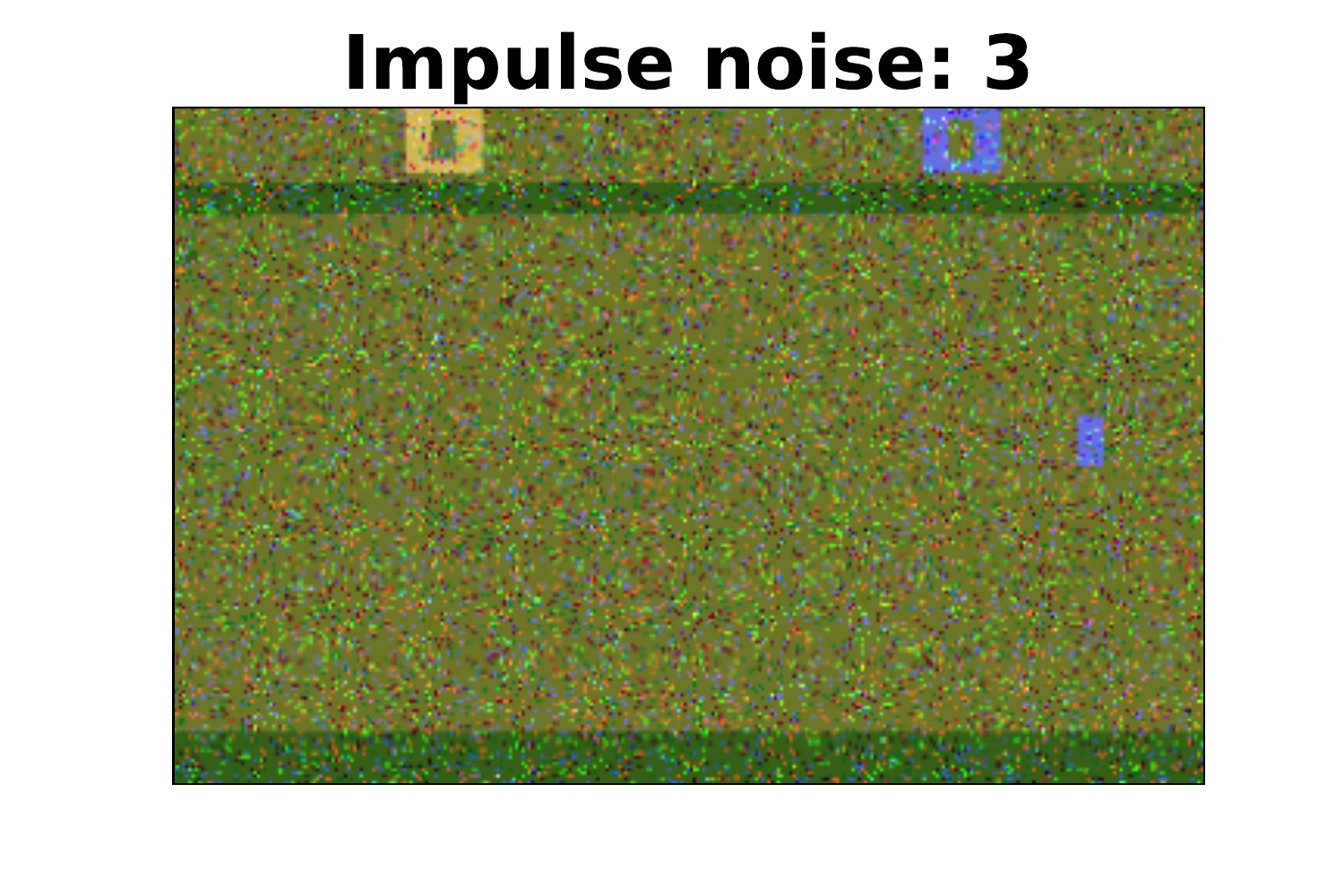}
        \caption{$p = 0.09$}
    \end{subfigure}
    \begin{subfigure}{0.24\textwidth}
        \includegraphics[width=\linewidth]{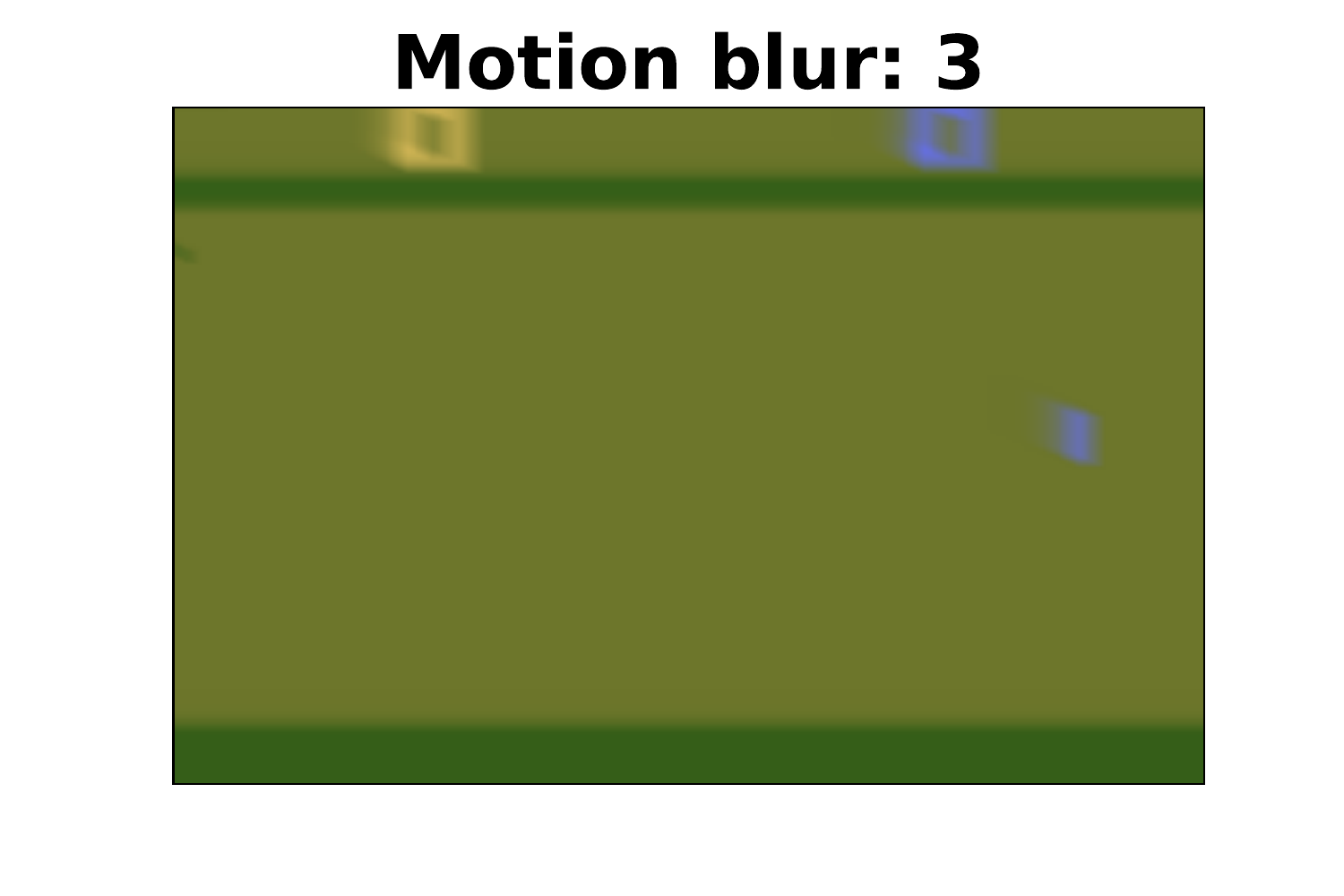}
        \caption{$\rho = 15\sigma = 8$}
    \end{subfigure}
    \begin{subfigure}{0.24\textwidth}
        \includegraphics[width=\linewidth]{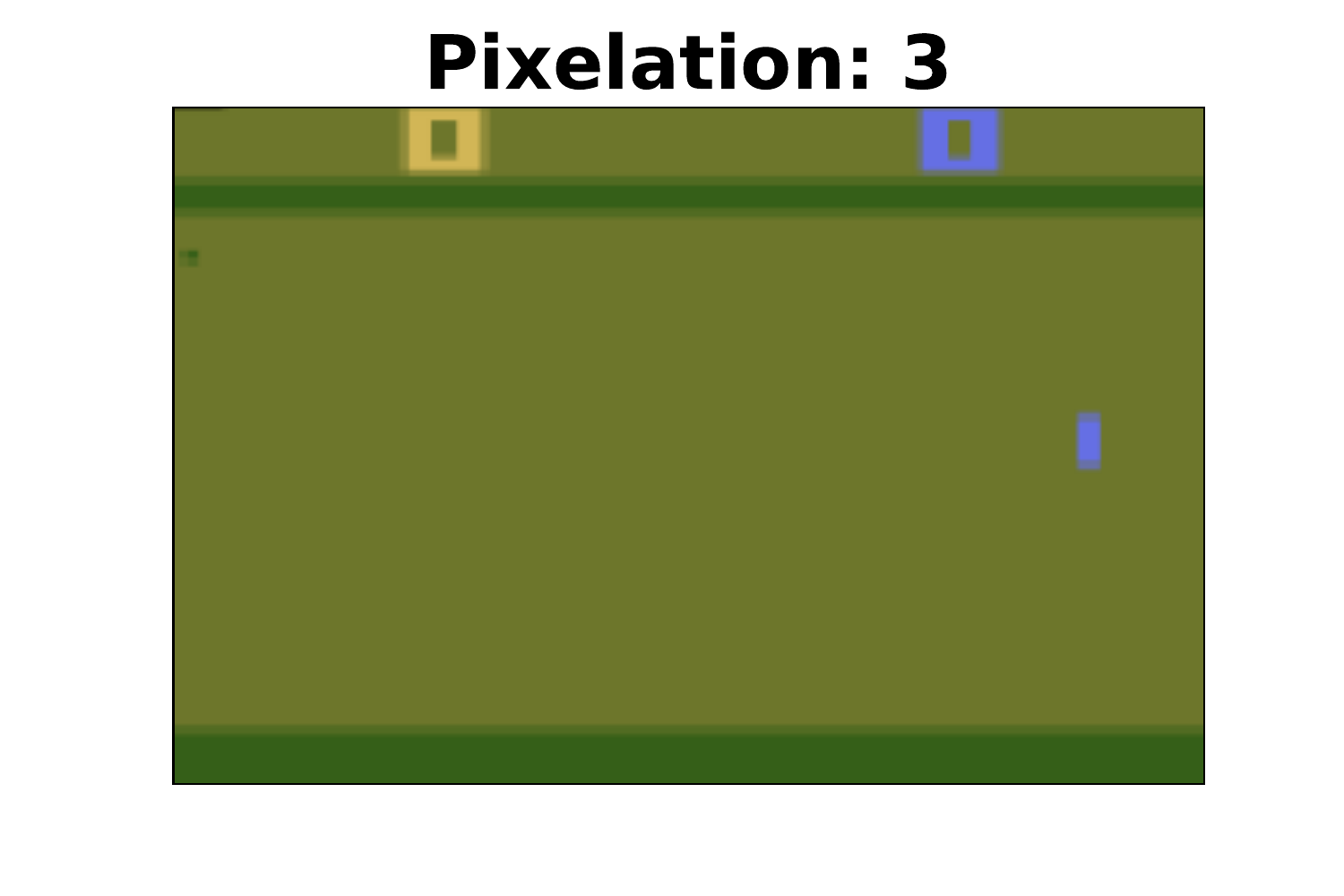}
        \caption{$f = 0.4$}
    \end{subfigure}

    \begin{subfigure}{0.24\textwidth}
        \includegraphics[width=\linewidth]{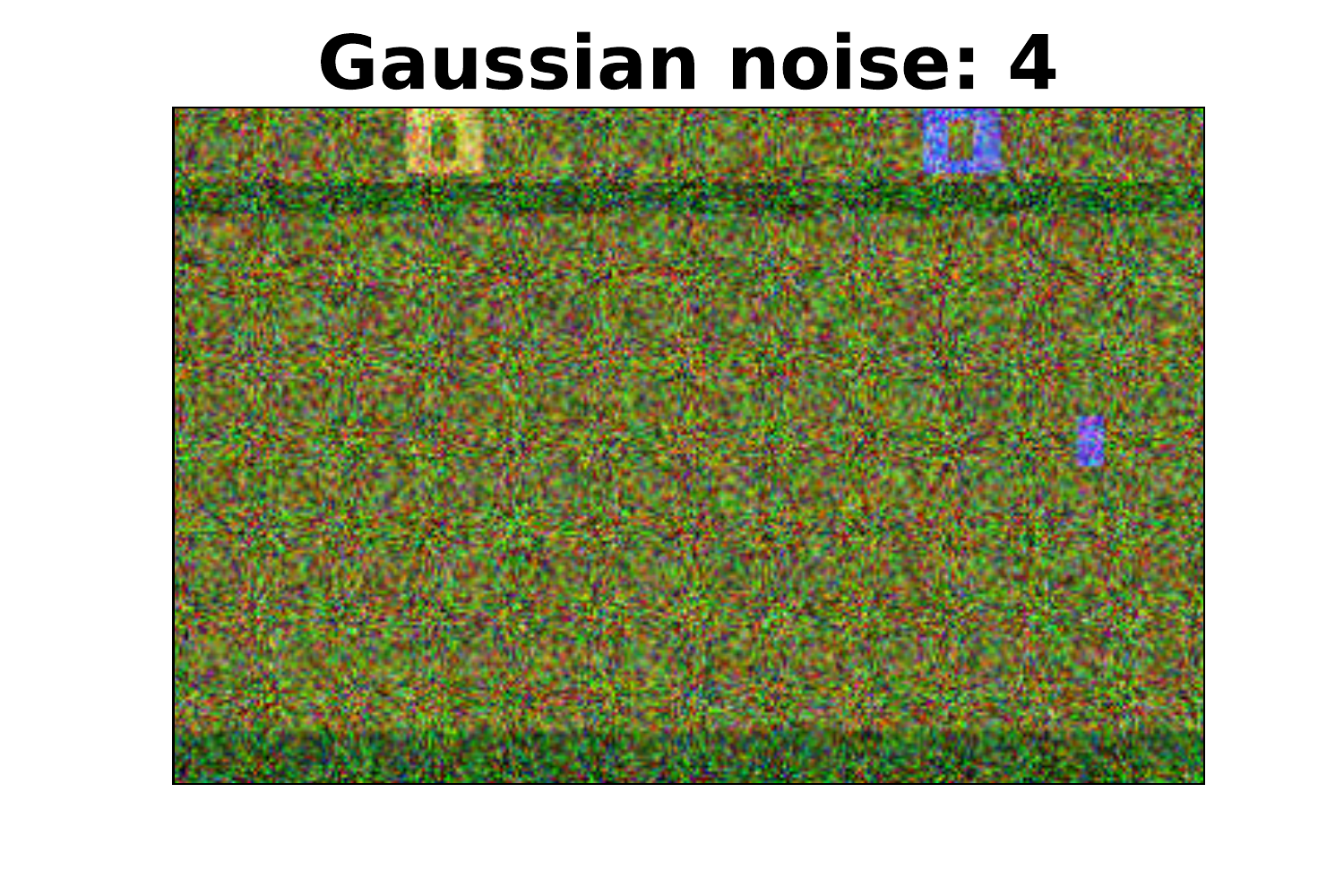}
        \caption{$\sigma = 0.26$}
    \end{subfigure}
    \begin{subfigure}{0.24\textwidth}
        \includegraphics[width=\linewidth]{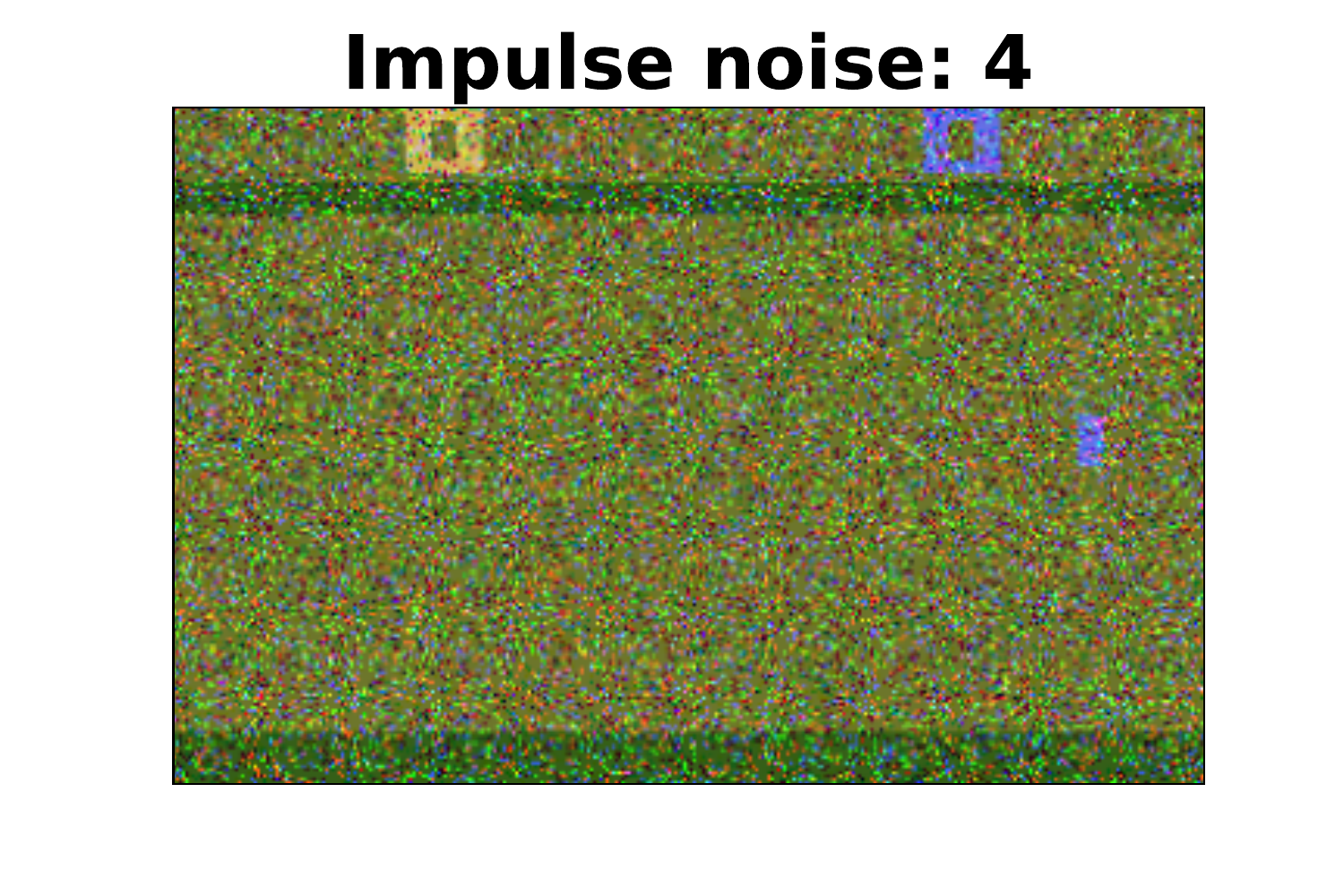}
        \caption{$p = 0.17$}
    \end{subfigure}
    \begin{subfigure}{0.24\textwidth}
        \includegraphics[width=\linewidth]{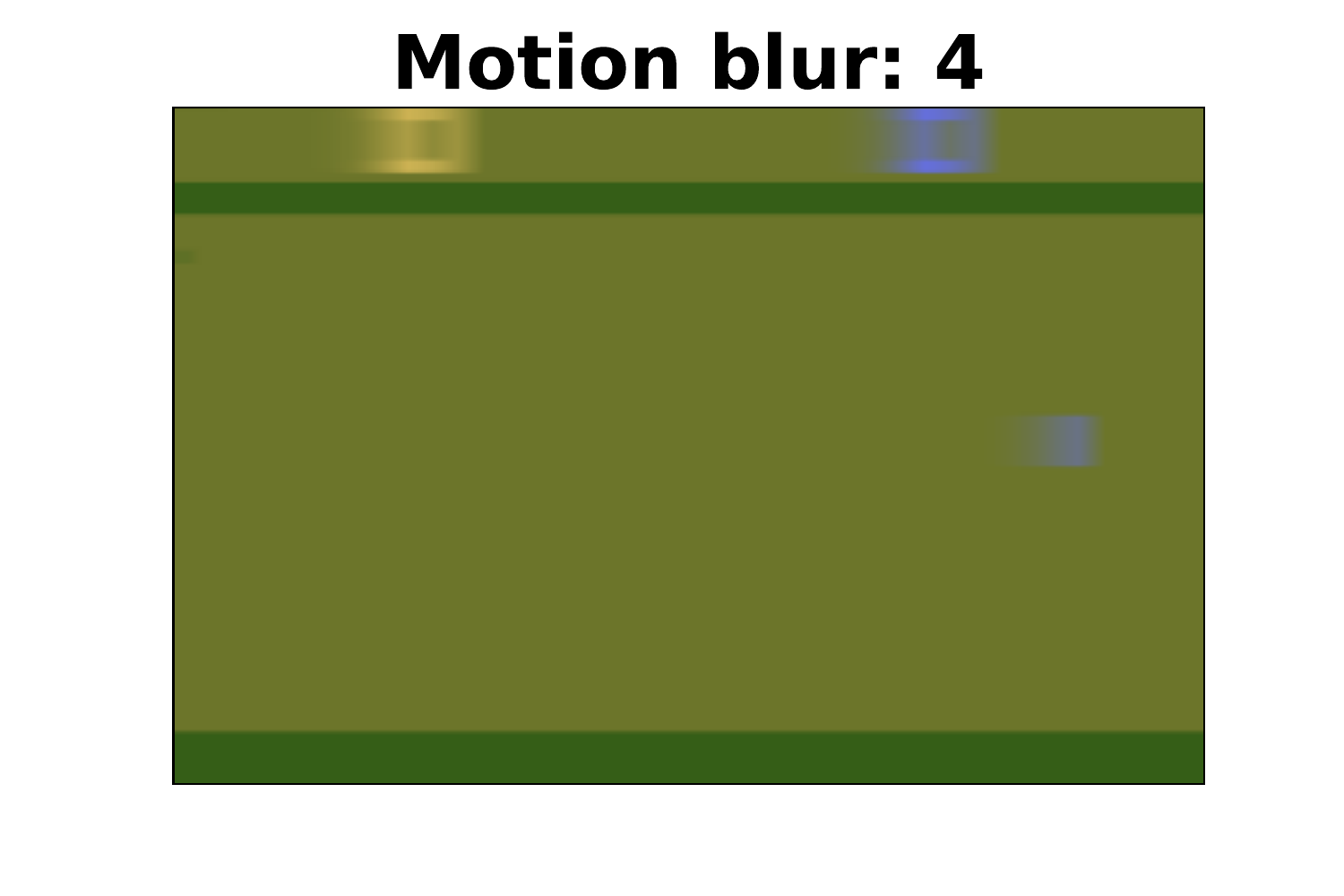}
        \caption{$\rho = 15\sigma = 12$}
    \end{subfigure}
    \begin{subfigure}{0.24\textwidth}
        \includegraphics[width=\linewidth]{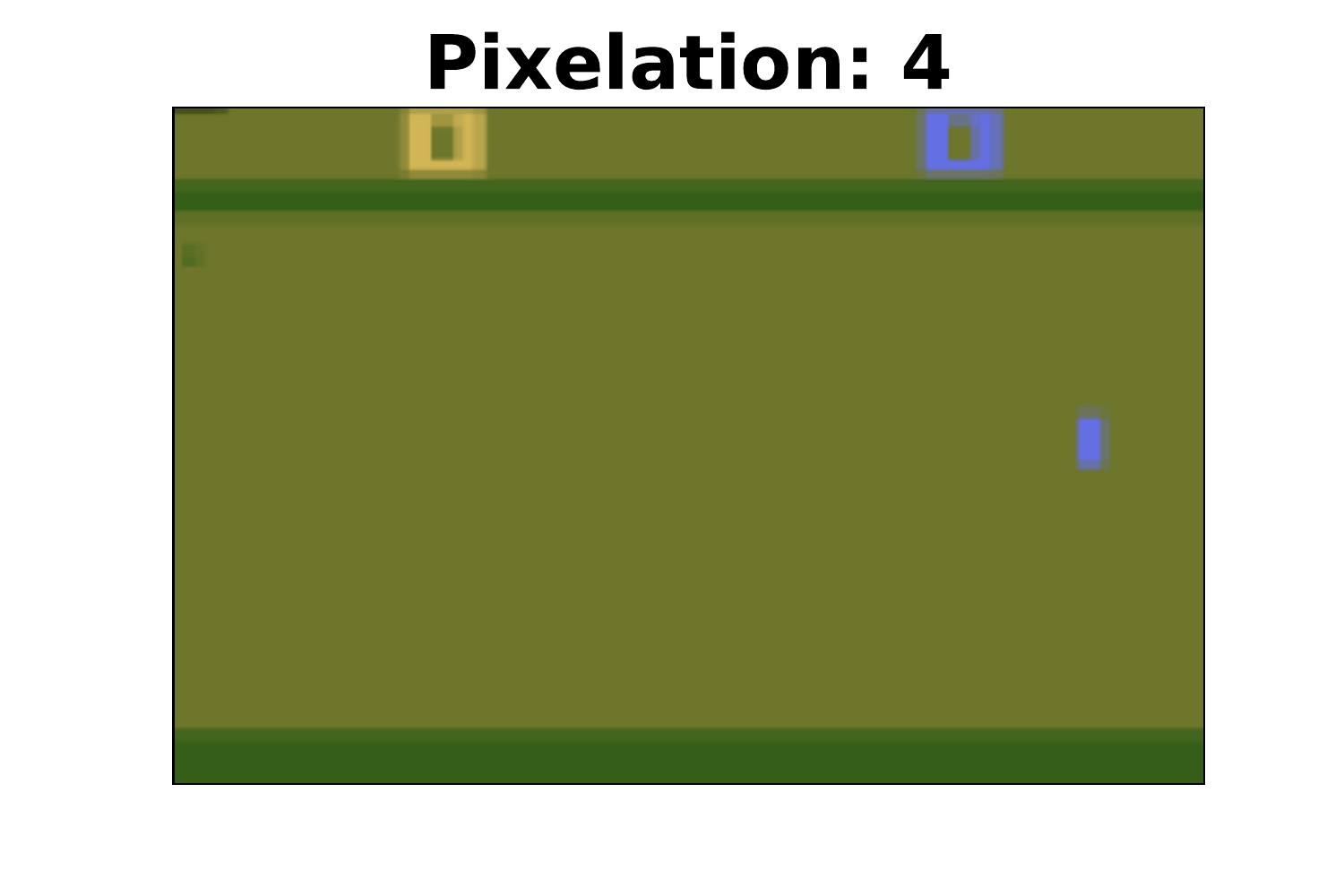}
        \caption{$f = 0.3$}
    \end{subfigure}

    \begin{subfigure}{0.24\textwidth}
        \includegraphics[width=\linewidth]{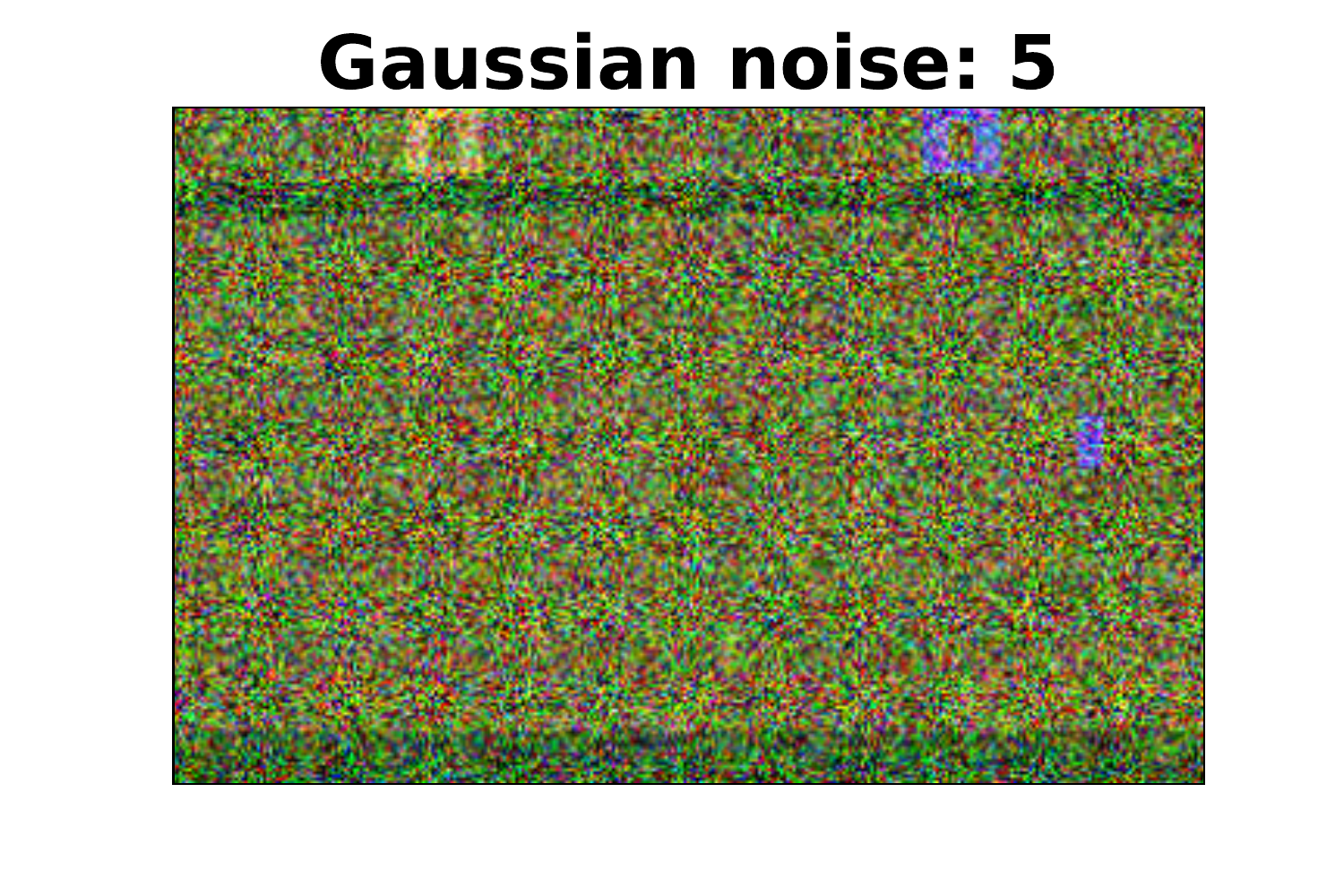}
        \caption{$\sigma = 0.38$}
    \end{subfigure}
    \begin{subfigure}{0.24\textwidth}
        \includegraphics[width=\linewidth]{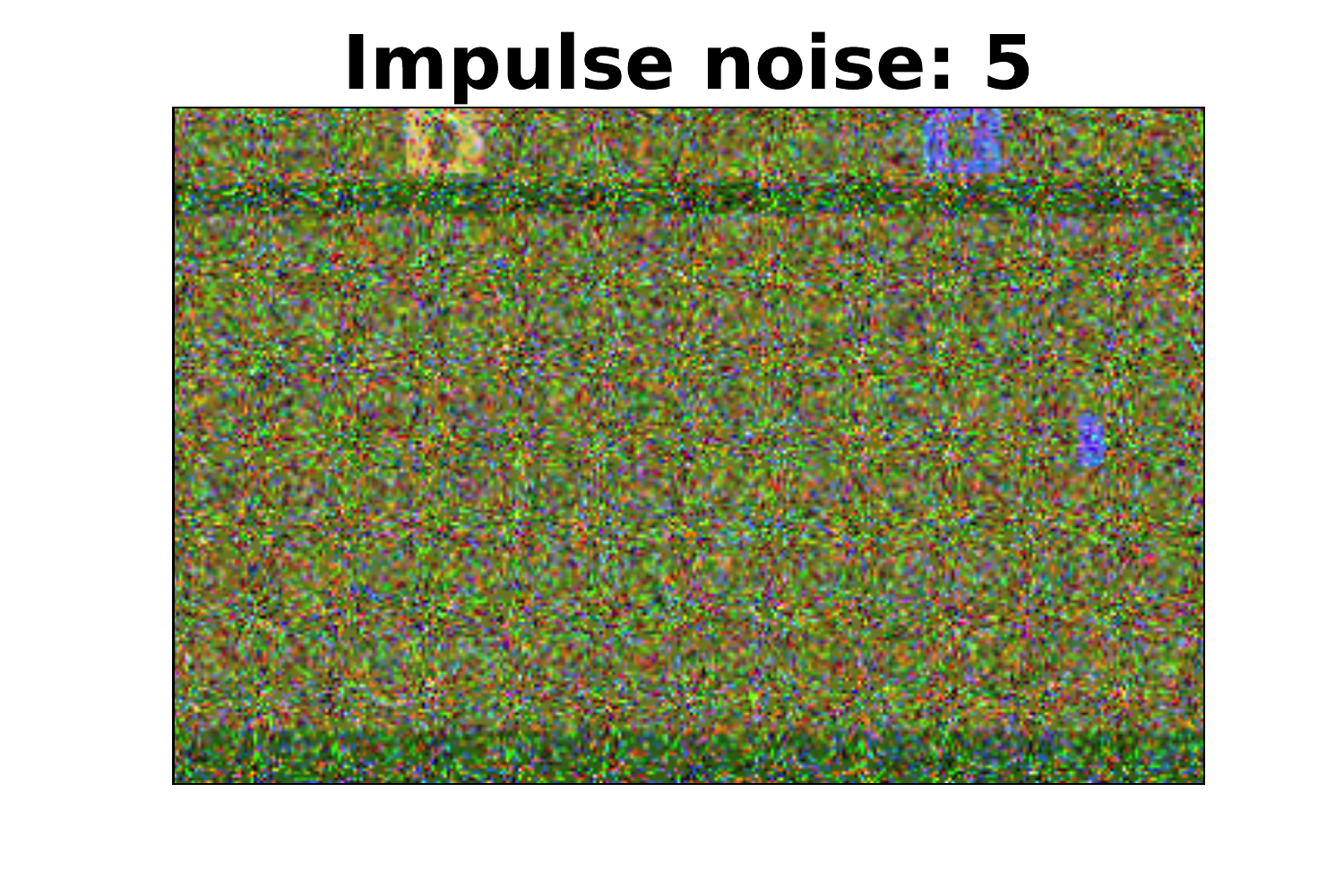}
        \caption{$p = 0.27$}
    \end{subfigure}
    \begin{subfigure}{0.24\textwidth}
        \includegraphics[width=\linewidth]{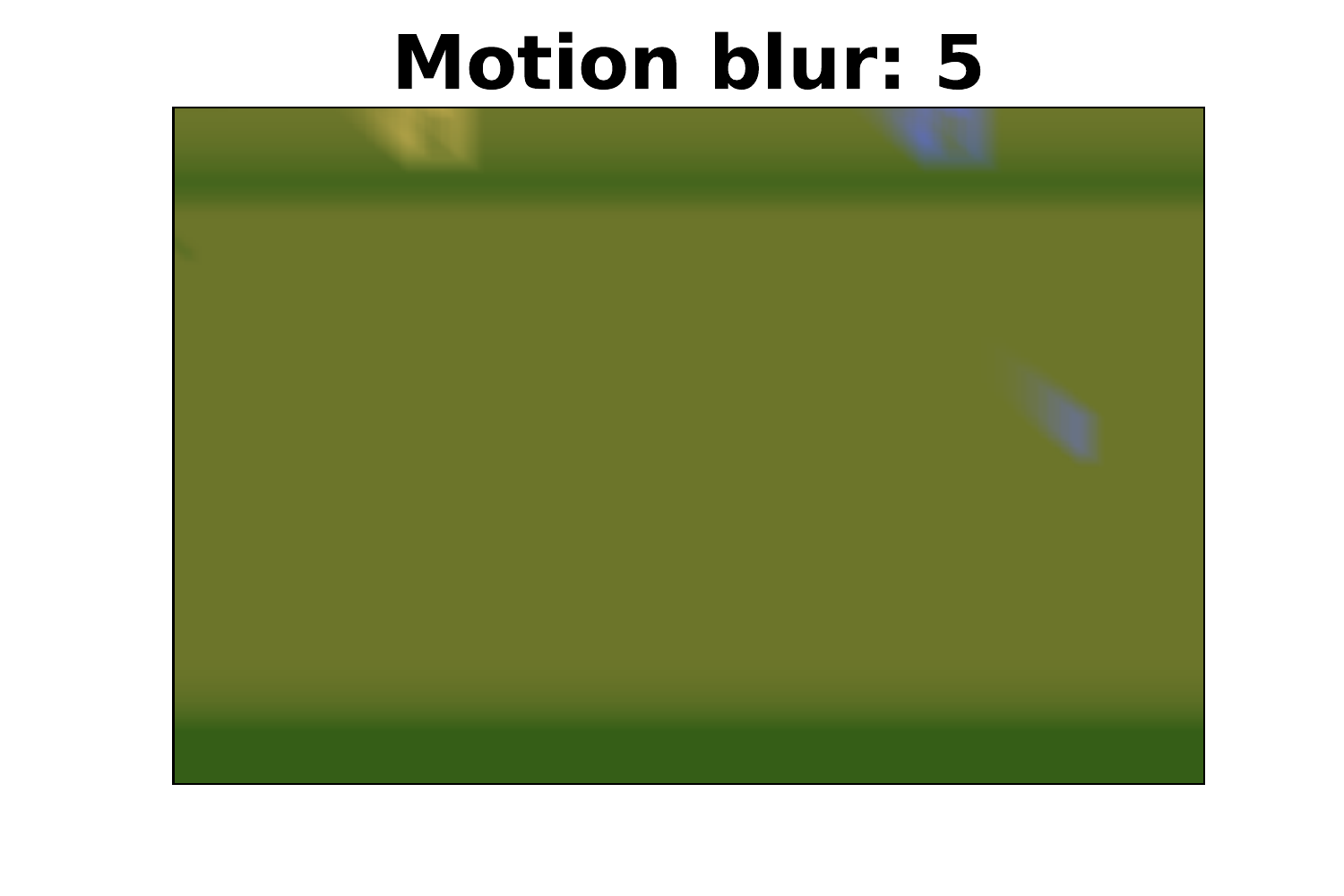}
        \caption{$\rho = 20\sigma = 15$}
    \end{subfigure}
    \begin{subfigure}{0.24\textwidth}
        \includegraphics[width=\linewidth]{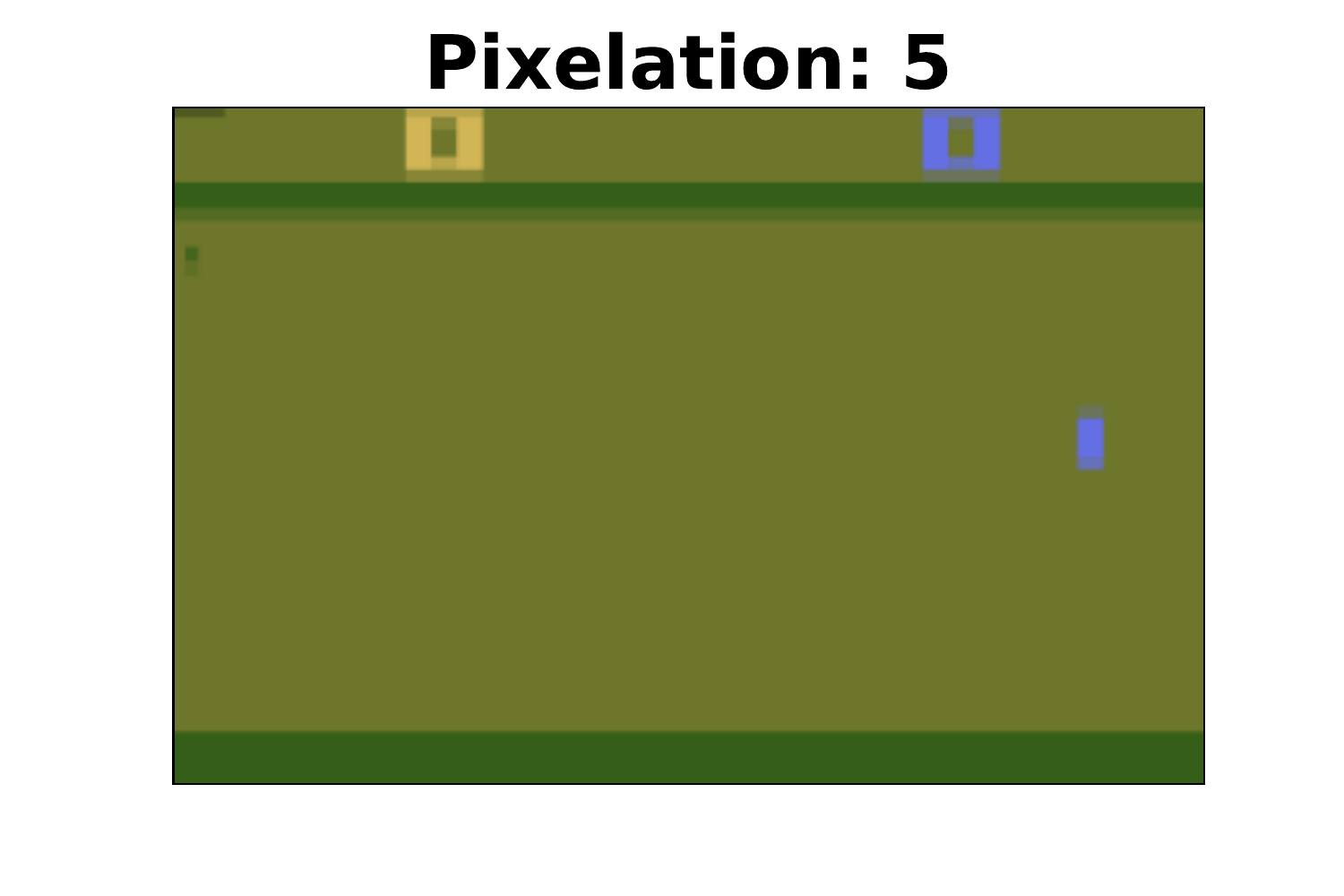}
        \caption{$f = 0.25$}
    \end{subfigure}
    \caption{Sample of different image/state observation corruptions and their variations along different severity parameters.}
    \label{image-corruptions}
\end{figure}

In this section we provide details of the state observation/image corruptions that we used to make OOD-Pong. Figure \ref{image-corruptions} makes a visual comparison of these corruptions across different parameters.

\textbf{Gaussian Noise}. Each pixel is added a sample drawn from a Gaussian distribution with mean zero and standard deviation $\sigma$.

\textbf{Impulse Noise}. Also known as salt and pepper noise. $p w h$ pixels in the image are randomly set to the minimum or maximum pixel value, where $p \in [0, 1]$ is the percentage of noisy pixels.

\textbf{Motion Blur}. A Gaussian blur kernel of size $(2\rho + 1) \times (2\rho + 1)$ and standard deviation $\sigma$ is generated, which is applied to the image at a random angle between $[-\frac{\pi}{4}, \frac{\pi}{4}]$ instead of being axis aligned.

\textbf{Pixelate}. The input image is downscaled to size $(f w, f h)$, and then upscaled to $(w, h)$. This process loses part of spatial information and the upscaling will make the image look pixelated. The parameter $f \in [0, 1]$ determines how much downscaling is performed.

\section{Detailed Experimental Results}
\FloatBarrier

In this section we provide all the experimental results for all environments, uncertainty methods, and RL algorithms, including multiple trials.

Table \ref{cartpole-dqn-dropout-ood-auc} shows the AUC scores obtained during multiple trials on custom versions of the cartpole environment using MC Dropout for uncertainty estimation.  MC Dropout produces a mean AUC score of around 0.71 along with a standard deviation of around 0.05 for the custom environment with a force of 2.5. It obtains a mean AUC score of around 0.69 along with a standard deviation of 0.07 for the custom environment with a gravity of 49. Similarly it achieves a mean AUC score of around 0.7 along with a standard deviation of 0.01 for the custom environment with the mass of the pole as 3. This shows that MC Dropout does a decent job in distinguishing between the in-distribution and out-of-distribution observations.  The low values of the standard deviation show that the OOD detection of MC Dropout is reproducible.

\begin{table}[h]
    \centering
    \begin{tabular}{@{}lccc@{}}
        \toprule
        & \multicolumn{1}{l}{Force: 2.5} & \multicolumn{1}{l}{Gravity: 49.0} & \multicolumn{1}{l}{Mass of the cart: 3.0} \\ \midrule
        AUC score trial 1 & 0.737                          &0.750     & 0.719             \\
        AUC score trial 2 & 0.688                          & 0.685     &0.691             \\
        AUC score trial 3 & 0.652                          & 0.606                             & 0.703                                     \\
        AUC score trial 4 & 0.780                          & 0.759                             & 0.708                                     \\
        AUC score trial 5 & 0.703                          & 0.626                             & 0.685                                     \\
        Mean AUC score $\pm$ Std    & \textbf{0.712 $\pm$ 0.049}                 & \textbf{0.685 $\pm$ 0.069}                    & \textbf{0.701 $\pm$ 0.013}                            \\ \bottomrule
    \end{tabular}
    \vspace*{0.2em}
    \caption{AUC scores obtained during multiple trials of OOD detection using MC Dropout on the custom versions of the cartpole environment along with their configurations. The mean AUC score and the standard deviation are also highlighted for each configuration.}
    \label{cartpole-dqn-dropout-ood-auc}
\end{table}

\begin{table}[t]
\resizebox{\textwidth}{!}{
\begin{tabular}{@{}lcccc@{}}
\toprule
Parameter         & \multicolumn{1}{l}{Force: 1.0} & \multicolumn{1}{l}{Gravity: 78.4} & \multicolumn{1}{l}{Length of the pole: 2} & \multicolumn{1}{l}{Mass of the cart: 9} \\ \midrule
AUC score trial 1 & 0.748                          & 0.844                             & 0.888                                     & 0.696                                   \\
AUC score trial 2 & 0.921                          & 0.834                             & 0.987                                     & 0.887                                   \\
AUC score trial 3 & 0.906                          & 0.758                             & 0.822                                     & 0.745                                   \\
AUC score trial 4 & 0.768                          & 0.693                             & 0.936                                     & 0.756                                   \\
AUC score trial 5 & 0.612                          & 0.992                             & 0.765                                     & 0.821                                   \\
Mean AUC score $\pm$ Std    & \textbf{0.791 $\pm$ 0.127}                 & \textbf{0.824 $\pm$ 0.112}                    & \textbf{0.880 $\pm$ 0.089}                            & \textbf{0.781 $\pm$ 0.074}                          \\ \bottomrule
\end{tabular}
}
\vspace*{0.2em}
\caption{AUC scores obtained during multiple trials of OOD detection using MC DropConnect on the custom versions of the cartpole environment along with their configurations. The mean AUC score and the standard deviation are also highlighted for each configuration.}
\label{cartpole-dqn-dropconnect-ood-auc}
\end{table}

Table \ref{cartpole-dqn-dropconnect-ood-auc} shows the AUC scores obtained during multiple trials of OOD detection on custom versions of the cartpole environment using MC DropConnect.  MC DropConnect produces a mean AUC score of around 0.79 along with  a standard deviation of around 0.13 on the custom environment with changed force of 1.0.  It achieves even higher mean AUC scores of around 0.82 with standard deviation of 0.11 and 0.88 with standard deviation of around 0.09 on the custom versions with changes in gravity and length of the pole respectively.  It also achieves a mean AUC score of around 0.78 along with  a standard deviation of around 0.07 on the custom environment with mass of the pole of 9. This shows that MC DropConnect does an excellent job in distinguishing between the ID and OOD observations. However, the standard deviations with MC DropConnect are slightly higher than that with MC Dropout. Overall, MC DropConnect does a better job in detecting OOD samples than MC Dropout but its performance is not always reproducible.

\begin{table}[t]
\centering
\begin{tabular}{@{}lcc@{}}
\toprule
Parameter         & \multicolumn{1}{l}{Gravity: 98} & \multicolumn{1}{l}{Length of the pole: 2} \\ \midrule
AUC score trial 1 & 0.756                             & 0.862                                     \\
AUC score trial 2 & 0.732                             & 0.826                                     \\
AUC score trial 3 & 0.734                             & 0.993                                     \\
AUC score trial 4 & 0.833                             & 0.864                                     \\
AUC score trial 5 & 0.803                             & 0.870                                     \\
Mean AUC score $\pm$ Std    & \textbf{0.772 $\pm$ 0.045}                    & \textbf{0.883 $\pm$ 0.064}                            \\ \bottomrule
\end{tabular}
\vspace*{0.2em}
\caption{AUC scores obtained during multiple trials of OOD detection using ensemble method on the custom versions of the cartpole environment along with their configurations. The mean AUC score and the standard deviation are also highlighted for each configuration.}
\label{cartpole-dqn-ensemble-ood-auc}
\end{table}

Table \ref{cartpole-dqn-ensemble-ood-auc} shows the AUC scores obtained during multiple trials on custom versions of the cartpole environment using ensemble for uncertainty estimation.  The ensemble method achieves a highest mean AUC score of around 0.88 with a standard deviation of around 0.06 for the custom version with the length of 2.  It also achieves a mean AUC score of around 0.77 with a standard deviation of around 0.05 for the custom version with gravity of 98. The higher AUC scores show that the ensemble method does an excellent job in distinguishing between the in-distribution and out-of-distribution observations.  The lower values of the standard deviation also show that the OOD detection using ensemble method is reproducible.

\begin{table}[t]
\centering
\begin{tabular}{@{}lccc@{}}
\toprule
                  & \multicolumn{1}{l}{Gravity: 50} & \multicolumn{1}{l}{Length of the pole: 5} & \multicolumn{1}{l}{Mass of the pole: 5} \\ \midrule
AUC score trial 1 & 0.726   & 0.569                                     & 0.674           \\
AUC score trial 2 & 0.626                           & 0.613                                     & 0.642                                   \\
AUC score trial 3 & 0.610                           & 0.578                                     & 0.726                                   \\
AUC score trial 4 & 0.618                           & 0.656                                     & 0.701                                   \\
AUC score trial 5 & 0.727                           & 0.651                                     & 0.677                                   \\
Mean AUC score $\pm$ Std   & \textbf{0.661 $\pm$ 0.060}                  & \textbf{0.614 $\pm$ 0.040}                            & \textbf{0.684 $\pm$ 0.031}                          \\ \bottomrule
\end{tabular}
\vspace*{0.2em}
\caption{AUC scores obtained during multiple trials of OOD detection using MC Dropout on the custom versions of the pendulum environment along with their configurations. The mean AUC score and the standard deviation are also highlighted for each configuration.}
\label{pendulum-ddpg-dropout-auc}
\end{table}

Table \ref{pendulum-ddpg-dropout-auc} shows the AUC scores for multiple trials of OOD detection using MC Dropout method with custom versions of the pendulum environment. MC Dropout achieves a mean AUC score of around 0.66 along with the standard deviation of 0.06 for the custom version where the gravity was increased to 50. The method achieves a mean AUC score of 0.614 along with a standard deviation of 0.04 for the custom environment with an increased length of 5. Similary, the method achieves a mean AUC score of 0.684 along with a standard deviation of 0.03 for the custom environment with an increased mass of 5. The smaller standard deviation values show that the performance of OOD detection by the dropout method is robust and reproducible.

\begin{table}[t]
\centering
\begin{tabular}{@{}lccc@{}}
\toprule
                  & \multicolumn{1}{l}{Gravity: 50} & \multicolumn{1}{l}{Length of the pole: 5} & \multicolumn{1}{l}{Mass of the pole: 5} \\ \midrule
AUC score trial 1 & 0.550                           & 0.628                                     & 0.533                                   \\
AUC score trial 2 & 0.602                           & 0.682                                     & 0.543                                   \\
AUC score trial 3 & 0.587                           & 0.555                                     & 0.529                                   \\
AUC score trial 4 & 0.599                           & 0.588                                     & 0.659                                   \\
AUC score trial 5 & 0.592                           & 0.726                                     & 0.715                                   \\
Mean AUC score $\pm$ Std   & \textbf{0.586 $\pm$ 0.021}                  & \textbf{0.636 $\pm$ 0.069}                            & \textbf{0.596 $\pm$ 0.086}                          \\ \bottomrule
\end{tabular}
\vspace*{0.2em}
\caption{AUC scores obtained during multiple trials of OOD detection using MC DropConnect  on the custom versions of the pendulum environment along with their configurations. The mean AUC score and the standard deviation are also highlighted for each configuration.}
\label{pendulum-ddpg-dropconnect-auc}
\end{table}

Table \ref{pendulum-ddpg-dropconnect-auc} shows the AUC scores for multiple trials of OOD detection using MC DropConnect  method with custom versions of the pendulum environment. MC DropConnect  achieves a mean AUC score of around 0.59 along with the standard deviation of 0.02 for the custom version where the gravity was increased to 50. The method achieves a mean AUC score of 0.636 along with a standard deviation of around 0.07 for the custom environment with an increased length of 5. Similarly, the method achieves a mean AUC score of 0.596 along with a standard deviation of around 0.09 for the custom environment with an increased mass of 5. The standard deviation values with MC DropConnect are slightly higher than those from the MC Dropout method. This shows that MC Dropout is more robust than MC DropConnect.

\begin{table}[t]
\centering
\begin{tabular}{@{}lcc@{}}
\toprule
                  & \multicolumn{1}{l}{Length of the pole: 0.1} & \multicolumn{1}{l}{Mass of the pole: 0.05} \\ \midrule
AUC score trial 1 & 0.583                                       & 0.582                                      \\
AUC score trial 2 & 0.619                                       & 0.546                                      \\
AUC score trial 3 & 0.612                                       & 0.587                                      \\
AUC score trial 4 & 0.601                                       & 0.596              \\
AUC score trial 5 & 0.592                                       & 0.556                                      \\
Mean AUC score $\pm$ Std    & \textbf{0.602 $\pm$ 0.014}                              & \textbf{0.573 $\pm$ 0.021}                             \\ \bottomrule
\end{tabular}
\vspace*{0.2em}
\caption{AUC scores obtained during multiple trials of OOD detection using ensemble on the custom versions of the pendulum environment along with their configurations. The mean AUC score and the standard deviation are also highlighted for each configuration.}
\label{pendulum-ddpg-ensemble-auc}
\end{table}

Table \ref{pendulum-ddpg-ensemble-auc} shows the AUC scores for multiple trials of OOD detection using the ensemble method with custom versions of the pendulum environment. The Ensemble method achieves a mean AUC score of around 0.6 along with the standard deviation of around 0.01 for the custom version where the length of the pole was decreased to 0.1. The method achieves a mean AUC score of 0.573 along with a standard deviation of around 0.021 for the custom environment with a decreased mass of 0.05. The standard deviation values are lower compared to that from MC DropConnect and MC Dropout methods. The results from the ensemble method show that it is more robust than the other two methods.

\begin{table}[t]
\centering
\begin{tabular}{@{}lccc@{}}
\toprule
                   & \multicolumn{3}{c}{Gaussian noise}                                                                  \\
                    & \multicolumn{1}{l}{$\sigma = 0.18$} & \multicolumn{1}{l}{$\sigma = 0.26$} & \multicolumn{1}{l}{$\sigma = 0.38$} \\ \midrule
AUC score trial 1  & 0.684                           & 0.738                           & 0.813                           \\
AUC score: trial 2 & 0.696                           & 0.747                           & 0.801                           \\
AUC score: trial 3 & 0.683                           & 0.749                           & 0.828                           \\
AUC score: trial 4 & 0.644                           & 0.737                           & 0.827                           \\
AUC score: trial 5 & 0.699                           & 0.749                           & 0.805                           \\
Mean AUC score $\pm$ Std    & \textbf{0.681 $\pm$ 0.022}                  & \textbf{0.744 $\pm$ 0.006}                  & \textbf{0.815 $\pm$ 0.012}                  \\ \bottomrule
\end{tabular}
\vspace*{0.2em}
\caption{AUC scores obtained during multiple trials of OOD detection using MC DropConnect on the custom versions with changing severity level in the gaussian noise for the pong environment. The mean AUC scores and the standard deviations are also highlighted.}
\label{pong-dropconnect-gaussian-auc}
\end{table}

Table \ref{pong-dropconnect-gaussian-auc} shows the AUC scores from different trials of OOD detection using the MC DropConnect method for the custom pong environment with changing severity levels in gaussian noise. The MC DropConnect method achieves a mean AUC score of 0.681 with a standard deviation of around 0.02 for a severity level of 3. It achieves higher mean scores of 0.744 with standard deviation of almost zero and 0.815 with standard deviation of almost 0.01 with severity levels of 4 and 5 respectively.  This shows that the the OOD detection performance of MC DropConnect increases with increase in the severity level in gaussian noise. The very low standard deviation values also highlight the reproducibility of the performance.

\begin{table}[t]
\centering
\begin{tabular}{@{}lccc@{}}
\toprule
                   & \multicolumn{3}{c}{Impulse noise}                                                                   \\
                   & \multicolumn{1}{l}{$p = 0.09$} & \multicolumn{1}{l}{$p = 0.17$} & \multicolumn{1}{l}{$p = 0.27$} \\ \midrule
AUC score trial 1  & 0.712                           & 0.750                           & 0.822                           \\
AUC score: trial 2 & 0.698                           & 0.766                           & 0.813                           \\
AUC score: trial 3 & 0.676                           & 0.740                           & 0.797                           \\
AUC score: trial 4 & 0.683                           & 0.775                           & 0.818                           \\
AUC score: trial 5 & 0.688                           & 0.752                           & 0.818                           \\
Mean AUC score  $\pm$ Std    & \textbf{0.691 $\pm$ 0.014}                  & \textbf{0.756 $\pm$ 0.014}                  & \textbf{0.814 $\pm$ 0.010}                  \\ \bottomrule
\end{tabular}
\vspace*{0.2em}
\caption{AUC scores obtained during multiple trials of OOD detection using MC DropConnect on the custom versions with changing severity level in the impulse noise for the pong environment. The mean AUC scores and the standard deviations are also highlighted.}
\label{pong-dropconnect-impulse-auc}
\end{table}

Table \ref{pong-dropconnect-impulse-auc} shows the AUC scores from different trials of OOD detection using the MC DropConnect method for the custom pong environment with changing severity levels in impulse noise. The MC DropConnect method achieves mean AUC scores of 0.69 with standard deviation of around 0.01 for severity level of 3. It achieves a mean AUC score  0.76 with a standard deviation of around 0.01 and 0.814 with a standard deviation of 0.01 for severity levels of 4 and 5 respectively. Similar to the gaussian noise, the OOD detection performance of MC DropConnect also increases with an increase in the corruption levels of impulse noise in the environment. The very low standard deviation values also highlight the reproducibility of the performance.

\begin{table}[!t]
\centering
\begin{tabular}{@{}lccc@{}}
\toprule
                   & \multicolumn{3}{c}{Motion blur}                                                                     \\
                   & \multicolumn{1}{l}{$\rho = 15, \sigma = 8$} & \multicolumn{1}{l}{$\rho = 15, \sigma = 12$} & \multicolumn{1}{l}{$\rho = 20, \sigma = 15$} \\ \midrule
AUC score trial 1  & 0.707                           & 0.663                           & 0.645                           \\
AUC score: trial 2 & 0.684                           & 0.662                           & 0.668                           \\
AUC score: trial 3 & 0.693                           & 0.682                           & 0.657                           \\
AUC score: trial 4 & 0.684                           & 0.677                           & 0.668                           \\
AUC score: trial 5 & 0.705                           & 0.656                           & 0.685                           \\
Mean AUC score $\pm$ Std     & \textbf{0.695 $\pm$ 0.011}                  & \textbf{0.668 $\pm$ 0.011}                  & \textbf{0.665 $\pm$ 0.015}                  \\ \bottomrule
\end{tabular}
\vspace*{0.2em}
\caption{AUC scores obtained during multiple trials of OOD detection using MC DropConnect on the custom versions with changing severity level in the motion blur for the pong environment. The mean AUC scores and the standard deviations are also highlighted.}
\label{pong-dropconnect-motion-auc}
\end{table}

Table \ref{pong-dropconnect-motion-auc} shows the AUC scores from different trials of OOD detection using the MC DropConnect method for the custom pong environment with changing severity levels in motion blur corruption. The MC DropConnect method achieves mean AUC scores of around 0.7 with a standard deviation of 0.01, for severity level of 3. It achieves a mean AUC score of 0.67 with a standard deviation of 0.01 and 0.665 with a standard deviation of around 0.02 for severity levels of 4 and 5 respectively.  Unlike the gaussian noise and the impulse noise, the OOD detection performance of MC DropConnect is not affected much by varying levels of motion blur in the environment. However, the lower standard deviation values highlight the reproducibility of its performance.

\begin{table}[!tb]
\centering
\begin{tabular}{@{}lcc@{}}
\toprule
                   & \multicolumn{2}{c}{Pixelate}                                      \\
                   & \multicolumn{1}{l}{$f = 0.4$} & \multicolumn{1}{l}{$f = 0.3$} \\ \midrule
AUC score trial 1  & 0.578                           & 0.556                           \\
AUC score: trial 2 & 0.598                           & 0.560                           \\
AUC score: trial 3 & 0.606                           & 0.553                           \\
AUC score: trial 4 & 0.584                           & 0.553                           \\
AUC score: trial 5 & 0.579                           & 0.565                           \\
Mean AUC score $\pm$ Std    & \textbf{0.589 $\pm$ 0.012}                  & \textbf{0.558 $\pm$ 0.005}                  \\ \bottomrule
\end{tabular}
\vspace*{0.2em}
\caption{AUC scores obtained during multiple trials of OOD detection using MC DropConnect on the custom versions with changing severity level in pixelation for the pong environment. The mean AUC scores and the standard deviations are also highlighted.}
\label{pong-dropconnect-pixelate-auc}
\end{table}

Table \ref{pong-dropconnect-pixelate-auc} shows the AUC scores from different trials of OOD detection using the MC DropConnect method for the custom pong environment with changing severity levels in pixelation. The MC DropConnect method achieves mean AUC scores of 0.589 with a standard deviation of around 0.01 and 0.558 with a standard deviation of almost zero for severity levels of 3 and 4 respectively.  The slightly lower values of the AUC scores show that the method is not very efficient in detecting the OOD observations due to corruption from pixelation. Also, the OOD detection performance is not affected much by varying levels of pixelation in the environment. However, the lower standard deviation values show that the OOD detection performance of the MC DropConnect is reproducible.

\begin{table}[tb]
\centering
\begin{tabular}{@{}lccc@{}}
\toprule
                   & \multicolumn{3}{c}{Gaussian noise}                                                                  \\
                   & \multicolumn{1}{l}{$\sigma = 0.18$} & \multicolumn{1}{l}{$\sigma = 0.26$} & \multicolumn{1}{l}{$\sigma = 0.38$}\\
                    \midrule
AUC score trial 1  & 0.647                           & 0.728                           & 0.819                           \\
AUC score: trial 2 & 0.621                           & 0.731                           & 0.823                           \\
AUC score: trial 3 & 0.634                           & 0.735                           & 0.829                           \\
AUC score: trial 4 & 0.627                           & 0.730                           & 0.831                           \\
AUC score: trial 5 & 0.644                           & 0.730                           & 0.821                           \\
Mean AUC score $\pm$ Std     & \textbf{0.635 $\pm$ 0.011}                  & \textbf{0.731 $\pm$ 0.003}                  & \textbf{0.825 $\pm$ 0.005}                  \\ \bottomrule
\end{tabular}
\vspace*{0.2em}
\caption{AUC scores obtained during multiple trials of OOD detection using ensemble on the custom versions with changing severity level in the gaussian noise for the pong environment. The mean AUC scores and the standard deviations are also highlighted.}
\label{pong-ensemble-gaussian-auc}
\end{table}

Table \ref{pong-ensemble-gaussian-auc} shows the AUC scores from different trials of OOD detection using the ensemble method for the custom pong environments with changing severity level in gaussian noise. The ensemble method achieves mean AUC scores of around 0.64 with a standard deviation of 0.01 with a severity level of 3. It achieves a mean AUC score of 0.731 with a standard deviation of almost zero and 0.825 with a standard deviation of almost zero for corruption levels of 4 and 5 respectively. This shows that the OOD detection performance of the ensemble method increases with the increase in the corruption level of gaussian noise in the environment. The lower standard deviation values also highlight the reproducibility of the performance. Overall, the ensemble method does an excellent job in detecting OOD observations due to gaussian noise.

\begin{table}[tb]
\centering
\begin{tabular}{@{}lccc@{}}
\toprule
                   & \multicolumn{3}{c}{Impulse noise}                                                                   \\
                   & \multicolumn{1}{l}{$p = 0.09$} & \multicolumn{1}{l}{$p = 0.17$} & \multicolumn{1}{l}{$p = 0.27$} \\ \midrule
AUC score trial 1  & 0.650                           & 0.758                           & 0.836                           \\
AUC score: trial 2 & 0.650                           & 0.769                           & 0.830                           \\
AUC score: trial 3 & 0.654                           & 0.755                           & 0.827                           \\
AUC score: trial 4 & 0.656                           & 0.763                           & 0.830                           \\
AUC score: trial 5 & 0.651                           & 0.748                           & 0.826                           \\
Mean AUC score $\pm$ Std    & \textbf{0.652 $\pm$ 0.002}                  & \textbf{0.759 $\pm$ 0.008}                  & \textbf{0.830 $\pm$ 0.004}                  \\ \bottomrule
\end{tabular}
\vspace*{0.2em}
\caption{AUC scores obtained during multiple trials of OOD detection using ensemble on the custom versions with changing severity level in the impulse noise for the pong environment. The mean AUC scores and the standard deviations are also highlighted.}
\label{pong-ensemble-impulse-auc}
\end{table}

Table \ref{pong-ensemble-impulse-auc} shows the AUC scores from different trials of OOD detection using the ensemble method for the custom pong environments with changing severity level in impulse noise. The ensemble method achieves mean AUC scores of around 0.65, 0.76 and 0.830 with a standard deviations of almost zero for severity levels of 3,  4 and 5 respectively. Similar to the gaussian noise, the OOD detection performance of the ensemble method also increase with an increase in the levels of impulse noise.  The lower standard deviation values highlight the reproducibility of the performance. Overall, the ensemble method does an excellent job in detecting OOD observations due to impulse noise.

\begin{table}[tb]
\centering
\begin{tabular}{@{}lccc@{}}
\toprule
                   & \multicolumn{3}{c}{Motion blur}                                                                     \\
                   & \multicolumn{1}{l}{$\rho = 15, \sigma = 8$} & \multicolumn{1}{l}{$\rho = 15, \sigma = 12$} & \multicolumn{1}{l}{$\rho = 20, \sigma = 15$} \\ \midrule
AUC score trial 1  & 0.886                           & 0.857                           & 0.831                           \\
AUC score: trial 2 & 0.884                           & 0.856                           & 0.822                           \\
AUC score: trial 3 & 0.902                           & 0.860                           & 0.841                           \\
AUC score: trial 4 & 0.904                           & 0.860                           & 0.838                           \\
AUC score: trial 5 & 0.910                           & 0.861                           & 0.839                           \\
Mean AUC score $\pm$ Std     & \textbf{0.897 $\pm$ 0.011}                  & \textbf{0.859 $\pm$ 0.002}                  & \textbf{0.834 $\pm$ 0.008}                  \\ \bottomrule
\end{tabular}
\vspace*{0.2em}
\caption{AUC scores obtained during multiple trials of OOD detection using ensemble on the custom versions with changing severity level in the motion blur for the pong environment. The mean AUC scores and the standard deviations are also highlighted.}
\label{pong-ensemble-motion-auc}
\end{table}

Table \ref{pong-ensemble-motion-auc} shows the AUC scores from different trials of OOD detection using the ensemble method for the custom pong environments with changing severity levels in motion blur corruption. The ensemble method achieves mean AUC scores of around 0.9 with a standard deviation of 0.01, 0.86 with a standard deviation of almost zero and 0.834 with a standard deviation of around 0.01 for severity levels of 3,  4 and 5 respectively. This shows that the MC DropConnect is very efficient in detecting the OOD observations due to corruption from motion blur. However, unlike for the gaussian noise and the impulse noise, the OOD detection performance is seen to be slightly affected adversely by the increase in levels of motion blur in the environment. But, the lower standard deviation values highlight the reproducibility of its performance. Overall, the higher AUC scores show that the ensemble method does an excellent job in detecting OOD observations due to motion blur.

\begin{table}[tb]
\centering
\begin{tabular}{@{}lcc@{}}
\toprule
                   & \multicolumn{2}{c}{Pixelate}                                      \\
                   & \multicolumn{1}{l}{$f = 0.4$} & \multicolumn{1}{l}{$f = 0.3$} \\ \midrule
AUC score trial 1  & 0.823                           & 0.634                           \\
AUC score: trial 2 & 0.823                           & 0.634                           \\
AUC score: trial 3 & 0.823                           & 0.634                           \\
AUC score: trial 4 & 0.823                           & 0.634                           \\
AUC score: trial 5 & 0.823                           & 0.634                           \\
Mean AUC score $\pm$ Std    & \textbf{0.823 $\pm$ 0.00}                  & \textbf{0.634 $\pm$ 0.00}                  \\ \bottomrule
\end{tabular}
\vspace*{0.2em}
\caption{AUC scores obtained during multiple trials of OOD detection using ensemble on the custom versions with changing severity level in pixelation for the pong environment. The mean AUC scores and the standard deviations are also highlighted.}
\label{pong-ensemble-pixelate-auc}
\end{table}

Table \ref{pong-ensemble-pixelate-auc} shows the AUC scores from different trials of OOD detection using the ensemble method for the custom pong environments with changing severity level in pixelation. The ensemble method achieves AUC scores of around 0.82 and 0.63 for severity levels of 3 and 4 respectively in all the trials. This shows that the ensemble is efficient in detecting the OOD observations due to corruption from pixelation especially when the corruption level is medium. Also, the OOD detection performance is adversely affected by the increase in corruption due to pixelation in the environment. However, constant AUC score across trials shows the reproducibility of OOD detection performance of the ensemble method.
\end{document}